\documentclass[sigconf]{acmart}

\usepackage{multirow}
\usepackage{natbib}
\renewcommand{\cite}{\citep}
\usepackage{longtable}
\usepackage{xcolor}
\usepackage{subcaption}
\usepackage{rotating}
\usepackage{graphicx}
\usepackage{booktabs}
\usepackage{hyperref}
\usepackage{float}
\usepackage{enumitem}
\usepackage{tikz}

\newcommand\citationnote{%
\footnotesize The accepted version is published in the Proceedings of the 2026 International Conference on Multimedia Retrieval (ICMR '26). DOI: \href{https://doi.org/10.1145/3805622.3810617}{https://doi.org/10.1145/3805622.3810617}.}
\newcommand\citationnoticebox{%
\begin{tikzpicture}[remember picture,overlay]
\node[anchor=north,yshift=-10pt] at (current page.north)
  {\fbox{\parbox{\dimexpr\textwidth-\fboxsep-\fboxrule\relax}{\citationnote}}};
\end{tikzpicture}%
}

\AtBeginDocument{%
  }

\setcopyright{acmlicensed}
\copyrightyear{2026}
\acmYear{2026}
\setcopyright{cc}
\setcctype{by}
\acmConference[ICMR '26]{International Conference on Multimedia Retrieval}{June 16--19, 2026}{Amsterdam, Netherlands}
\acmBooktitle{International Conference on Multimedia Retrieval (ICMR '26), June 16--19, 2026, Amsterdam, Netherlands}
\acmDOI{10.1145/3805622.3810617}
\acmISBN{979-8-4007-2617-0/2026/06}

\begin{document}

\title{Context-Aware Interpretable Representations for Retrieval and Graph Convolutional Network Classification}

%
\author{Thiago César Castilho Almeida}
\orcid{0000-0002-2167-0463}
\affiliation{%
    \institution{State University of São Paulo (UNESP)}
  \city{Rio Claro}
  \state{São Paulo}
  \country{Brazil}
}
\email{tc.almeida@unesp.br}

\author{Gustavo Rosseto Letício}
\orcid{0009-0008-3715-8991}
\affiliation{%
    \institution{State University of São Paulo (UNESP)}
  \city{Rio Claro}
  \state{São Paulo}
  \country{Brazil}
}
\email{gustavo.leticio@unesp.br}

\author{Vinicius Atsushi Sato Kawai}
\orcid{0000-0003-0153-7910}
\affiliation{%
    \institution{State University of São Paulo (UNESP)}
  \city{Rio Claro}
  \state{São Paulo}
  \country{Brazil}
}
\email{vinicius.kawai@unesp.br}

\author{Daniel Carlos Guimarães Pedronette}
\orcid{0000-0002-2867-4838}
\affiliation{%
    \institution{State University of São Paulo (UNESP)}
  \city{Rio Claro}
  \state{São Paulo}
  \country{Brazil}
}
\email{daniel.pedronette@unesp.br}

\renewcommand{\shortauthors}{Almeida, Letício, Kawai, and Pedronette}

\begin{abstract}
The advances in visual information modeling and representation during the last decades are remarkable, mainly supported by Convolutional Neural Networks, Transformer-based, and Foundation Models. Despite this progress, critical challenges regarding the nature of similarity assessment and model transparency have been neglected. A primary concern is the \textit{Geometric Gap}, where traditional pairwise measures fail to capture the intrinsic geometry of the dataset manifold. Furthermore, the \textit{Interpretability Gap} persists, as representations often lack alignment with human cognition. Therefore, how to provide interpretability to representations while maintaining low dimensionality and high effectiveness in downstream tasks remains an open challenge.
In this paper, we propose a novel unsupervised framework that integrates Manifold Learning strategies with Rank-based Interpretable Graph Embeddings. Our approach effectively bridges these gaps by first characterizing the contextual information of the dataset through manifold analysis and subsequently generating sparse, self-explainable embeddings. The proposed approach employs a flexible formulation, allowing different Manifold Learning and Representation Learning strategies. Extensive experimental evaluation across diverse datasets and features demonstrates that our Context-Aware representations not only provide intrinsic interpretability and dimensionality reduction but also maintain or enhance effectiveness in downstream tasks, specifically in image retrieval and semi-supervised classification using Graph Convolutional Networks (GCNs).
\end{abstract}



\maketitle
\citationnoticebox

\section{Introduction}

The fundamental pillars of Information Retrieval (IR) and Machine Learning (ML) reside in two interconnected components: data representation strategies and mechanisms for similarity assessment. In recent decades, multimedia retrieval has undergone a profound transformation driven by Deep Learning. The emergence of Convolutional Neural Networks (CNNs) \cite{he_deep_2016, zheng2017siftmeetscnndecade} and Vision Transformers (ViTs) \cite{vaswani_attention_2023, dosovitskiy_image_2021} has bridged the ``semantic gap'', mapping raw data into dense latent spaces where complex patterns are encoded. However, a critical asymmetry persists: while representation learning has advanced rapidly, the methods used to compare these representations have not evolved at the same pace.

Despite the non-linear nature of modern latent spaces, similarity assessment remains predominantly grounded in simple pairwise measures like Euclidean distance. These metrics suffer from the ``curse of dimensionality'' \cite{huang_review_2019} and fail to capture the intrinsic curvature of the data, leading to a mismatch known as the \textit{Geometric Gap} \cite{ermolov_hyperbolic_2022, levada_curvature_2022}. Furthermore, the individual dimensions of these dense representations lack explicit semantic meaning. This opacity creates a secondary barrier, the \textit{Interpretability Gap} \cite{anand2022explainable, zhao2021towards}, which prevents the adoption of retrieval systems in high-stakes domains where trust and human understanding are paramount \cite{rudin_stop_2019, marconato2023interpretability}. Consequently, the field faces a dual burden: optimizing for retrieval accuracy while ensuring the feature space remains semantically coherent to human observers.

Existing approaches address these gaps in isolation: interpretability methods such as Disentangled Representation Learning \cite{wang2024disentangled, higgins2017beta} and self-explaining models \cite{koh2020conceptbottleneckmodels, chen2019protopnet} either impose restrictive assumptions or require expensive supervision, while graph-based methods \cite{kipf_semi-supervised_2017} that capture manifold structure depend critically on neighborhood quality, which Euclidean metrics may fail to guarantee in high dimensions \cite{levada_curvature_2022}. Even recent rank-based interpretable methods \cite{fernando_rade_2020, almeida_grace_2025} that align embedding dimensions with semantic prototypes inherit this geometric flaw by relying on noisy initial neighborhoods.

In this paper, we propose a novel approach that integrates the robustness of manifold learning with the transparency of rank-based embeddings. We argue that ranking structures provide the optimal framework for encoding topology, as they are inherently resilient to the scale variances of raw feature spaces. By leveraging manifold learning to ``unfold'' the data geometry before graph construction, we ensure that the subsequent prototype selection is topologically accurate. This generates vector representations where each dimension corresponds to a highly effective prototype, simultaneously bridging the \textit{Geometric Gap} via context-aware processing and the \textit{Interpretability Gap} via self-explanatory dimensions.

The main contributions of this work are summarized as follows:
\begin{itemize}[noitemsep, topsep=4pt]
    \item \textit{Context-Aware and Interpretable Rank-based Framework:} We introduce a modular unsupervised formulation that exploits contextual ranking information through manifold learning to correct data geometry before embedding, allowing for the integration of different manifold learning algorithms and different rank-based representation learning strategies.
    \item \textit{Low-Dimensional and Effective Representation:} We demonstrate that our approach maintains or surpasses the effectiveness of high-dimensional features while significantly reducing dimensionality and providing semantic interpretability to each dimension.
    \item \textit{Representational Robustness:} The proposed approach is evaluated across multiple tasks, including content-based image retrieval and semi-supervised classification, proving its robustness as a general-purpose representation.
\end{itemize}

Experimental evaluations on diverse public datasets, using state-of-the-art CNN and Transformer backbones, indicate that the proposed method outperforms original features while offering qualitative insights.

\section{Related Work}
\label{sec:related_work}

This section reviews existing literature across three key areas: Network Representation Learning (NRL), which focuses on embedding graph structures; Disentangled Representation Learning (DRL), which attempts to isolate independent factors of variation; and Interpretable-by-Design Models, specifically those employing prototype-based reasoning.

\subsection{Network Representation Learning}
\label{subsec:nrl_methods}

Network Representation Learning (NRL) maps nodes from a high-dimensional graph structure into a low-dimensional vector space while preserving topological properties \cite{Chen_2020_graphrepresentationlearningasurvey}. Early ``shallow embedding'' approaches, inspired by Natural Language Processing, such as \textbf{DeepWalk} \cite{perozzi_deepwalk_2014} and \textbf{Node2vec} \cite{grover_node2vec_2016}, treat random walks on graphs as sentences, optimizing for node co-occurrence. Concurrently, methods like \textbf{LINE} \cite{tang_line_2015} focused on preserving explicit first- and second-order proximities.

The field subsequently shifted toward \textbf{Graph Neural Networks (GNNs)}, which integrate node attributes with structural information. Architectures like \textbf{GCNs} \cite{kipf_semi-supervised_2017} and \textbf{GATs} \cite{salehi2019graphattentionautoencoders} utilize message-passing mechanisms to aggregate neighbor information. To handle unlabeled data, unsupervised frameworks such as \textbf{GraphMAE} \cite{hou2022graphmaeselfsupervisedmaskedgraph} and \textbf{DGI} \cite{veličković2018deepgraphinfomax} have been developed to maximize mutual information or reconstruct masked features.

However, a major limitation of standard GNNs is that they produce ``black-box'' representations where dimensions are semantically opaque. To address this, \textit{Interpretable Graph Embedding} methods have emerged. Post-hoc approaches like \textbf{DINE} \cite{piaggesi_dine_2024} decompose existing embeddings into sparse, interpretable vectors. More recently, rank-based methods like \textbf{RaDE} \cite{fernando_rade_2020,fernando_rade_2022} and \textbf{GRaCE} \cite{almeida_grace_2025} generate interpretable embeddings directly from the graph structure. Unlike traditional GNNs, these approaches construct dimensions that explicitly encode similarity to specific prototype nodes, ensuring semantic alignment. While effective, these methods typically rely on input graphs constructed via standard distance metrics, leaving them vulnerable to the \textit{Geometric Gap} if the initial topology is not sufficiently refined.

\subsection{Disentangled Representation Learning}
\label{subsec:drl}

A primary goal of representation learning is to separate the underlying explanatory factors of data, a concept known as Disentangled Representation Learning (DRL) \cite{bengio2014representationlearningreviewnew}. In Computer Vision, Variational Autoencoders (VAEs) like $\boldsymbol \beta$\textbf{-VAE} \cite{higgins2017beta} enforce statistical independence among latent dimensions. These principles have been adapted to graphs by methods like \textbf{DisenGCN} \cite{ma2019disengcn} and \textbf{DiSeNE} \cite{piaggesi2025disentangled}, which decompose input graphs into independent factor graphs.

However, applying strict disentanglement to general-purpose retrieval is problematic. As noted in \cite{locatello2019challenging}, unsupervised disentanglement is theoretically impossible without strong inductive biases. Furthermore, DRL relies on the assumption that semantic factors are statistically independent. In complex multimedia manifolds, concepts are naturally correlated (e.g., ``ocean'' and ``blue''). Forcing independence can distort the intrinsic geometry of the data. 

In contrast, rank-based strategies adopt a philosophy of \textit{representational coverage} rather than statistical orthogonality. Instead of forcing latent dimensions to be independent factors, methods like RaDE and GRaCE select \textit{complementary} prototypes that maximize the coverage of the data manifold. This approach respects natural correlations, offering a more geometrically faithful form of interpretability where dimensions represent tangible exemplars rather than abstract, independent factors.

\subsection{Interpretable-by-Design Models}
\label{subsec:interpretable_models}

To avoid the unreliability of post-hoc explainers, recent research favors models that are interpretable by design. \textbf{Concept Bottleneck Models (CBMs)} \cite{koh2020conceptbottleneckmodels} map inputs to a set of human-understandable attributes (e.g., ``wing color'') before making a prediction. While recent works leverage Large Language Models (LLMs) to automate concept discovery \cite{oikarinen2023labelfreecbm, yang2023LaBo}, these methods suffer from the \textit{Linguistic Gap}, assuming all discriminative visual features can be verbally described, limiting their utility in specialized or abstract domains.

To bypass the linguistic bottleneck, \textbf{Prototype-based Models} adopt a \textit{this looks like that} paradigm. Architectures like \textbf{ProtoPNet} \cite{chen2019protopnet} and \textbf{TesNet} \cite{wang2021tesnet} learn prototypes as specific image patches. While transparent, they require specialized training pipelines and are difficult to adapt to pre-trained backbones \cite{gautam2024prototypical}.

Recent ``training-free'' approaches, such as \textbf{KMEx} \cite{gautam2024prototypical} and \textbf{IDEAL} \cite{ANGELOV2025IDEAL}, attempt to identify prototypes in the frozen latent space of pre-trained models using clustering or density peaks. However, a critical limitation persists: these methods rely on the assumption that the latent space is Euclidean. As established, high-dimensional spaces often exhibit non-Euclidean manifold structures \cite{levada_curvature_2022}. Relying on simple distances in such distorted spaces leads to the \textit{Geometric Gap}, where selected prototypes may not be truly representative. This motivates our proposed approach, which integrates manifold learning to unfold the data geometry before prototype selection, ensuring that explanations are topologically accurate.
    
\section{Proposed Approach}
\label{sec:proposed_approach}

\begin{figure*}[ht!]
    \centering
    \vspace{-3mm}
    \includegraphics[width=0.75\textwidth]{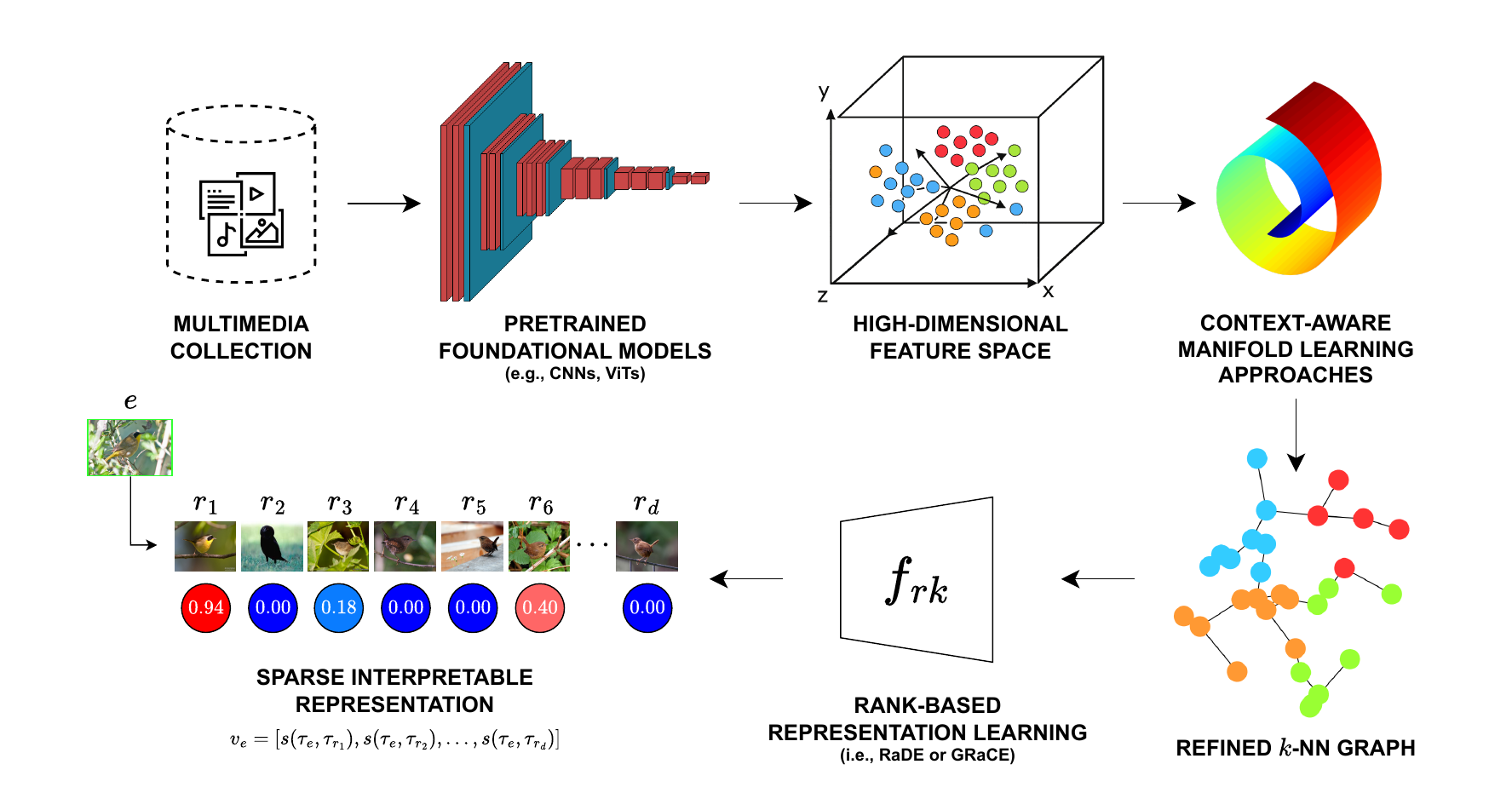}
    \vspace{-3mm}
    \caption{Overview of the proposed \textit{Context-Aware and Interpretable Rank-based Framework}.}
    \label{fig:proposed_approach}
\end{figure*}

We propose a fully unsupervised, modular framework that integrates manifold learning with rank-based interpretable embeddings to simultaneously address the \textit{Geometric} and \textit{Interpretability} gaps. We postulate that a ranked list serves as a robust regional discretization of the feature space. By leveraging this perspective, our framework transforms raw features into a sparse, interpretable embedding where context is encoded directly into the vector dimensions. The overall pipeline, illustrated in Figure \ref{fig:proposed_approach}, proceeds in three sequential stages: (1) Manifold Refinement, (2) Prototype Discovery, and (3) Interpretable Embedding Generation.

\subsection{Problem Formulation}
\label{subsec:problem_formulation}

Let $\mathcal{C} = \{e_1, e_2, \dots, e_N\}$ be a multimedia collection. A standard encoder $f_\theta$ maps each object $e_i$ to a feature vector $\mathbf{x}_i \in \mathbb{R}^D$. Conventionally, similarity is estimated via a pairwise distance function $\rho(\cdot, \cdot)$ (e.g., Euclidean distance), inducing a ranked list $\tau_q$ for each query $e_q$, such that:
\vspace{-2mm}
\begin{equation}
    \tau_q(i) < \tau_q(j) \Longleftrightarrow \rho(e_q, e_i) \le \rho(e_q, e_j).
\end{equation}

However, due to the non-isotropic nature of deep feature spaces, $\rho$ often fails to reflect true semantic affinity, introducing false positives in the neighborhood $\mathcal{N}(q, k)$.
Our objective is to learn a transformation $\Phi: \mathcal{X} \to \mathbb{R}^d$, $d \ll D$, mapping $\mathbf{x}_i$ to a new embedding $v_i$ that satisfies two conditions:
\begin{enumerate}
    \item \textbf{Topological Fidelity:} The similarity in the new space must reflect geodesic distances on the true data manifold $\mathcal{M}$, correcting the \textit{Geometric Gap}.
    \item \textbf{Intrinsic Interpretability:} The target space must be defined by a set of prototypes $\mathcal{R} = \{r_1, \dots, r_d\} \subset \mathcal{C}$, such that the $i$-th dimension of $v_e$ explicitly quantifies the similarity to the $i$-th prototype:
    \vspace{-2mm}
    \begin{equation}
        v_{e,i} = s(\tau_e, \tau_{r_i}).
    \end{equation}
\end{enumerate}

\subsection{The Framework}
\label{subsec:methodology}

The proposed methodology, detailed below, is modular, integrating Manifold Learning with Rank-based Graph Embeddings.

\subsubsection*{\textbf{Stage 1: Contextual Manifold Refinement}}
\label{subsubsec:stage1}

The first stage aims to ``unfold'' the data geometry to correct the \textit{Geometric Gap}. Raw feature spaces often suffer from the ``curse of dimensionality'' \cite{huang_review_2019}, creating short-circuit edges between semantically unrelated points. To mitigate this, we employ Manifold Learning strategies that exploit local neighborhood structures. We considered both Neighbor Embedding Projection and Context-Aware Similarity Learning \cite{pereiraferrero2024unsupervised} methods. Specifically, we adopt the following methods:

 $\bullet$ \textit{Uniform Manifold Approximation and Projection \linebreak (UMAP)~\cite{mcinnes_umap_2020}} operates as a neighbor embedding projection technique, constructing a weighted neighborhood graph in the original feature space and optimizing a low-dimensional representation that preserves local topological structures. This new representation is then used to compute the similarity relationships that better reflect the manifold information.

 $\bullet$ \textit{Rank-Based Diffusion Process for Assured Convergence (RDPAC)~\cite{pedronette_2021_rdpac}} follows a rank-based diffusion strategy, propagating affinity scores over a mutual neighborhood graph to smooth similarity estimates, attenuate noise, and ensure convergence, thereby producing refined ranked lists that exploit the manifold geometry.

 $\bullet$ \textit{Log-based Hypergraph of Ranking References \linebreak (LHRR)~\cite{pedronette_2019_lhrr}} models contextual similarity using a hypergraph structure, where relationships are derived from shared ranking references, with logarithmic weighting emphasizing top-$k$ neighbors.

 $\bullet$ \textit{Breadth-First Search Tree (BFSTREE)~\cite{PEDRONETTE2021_bfstree}} computes a new similarity measure by exploiting the manifold information provided by the ranking references in a breadth-first tree, thus computing refined similarity rankings.

Formally, this stage transforms the initial noisy rankings into a set of \textbf{refined ranked lists} $\mathcal{T}' = \{\tau'_1, \dots, \tau'_N\}$. In $\mathcal{T}'$, the rank position $\tau'_i(j)$ approximates the geodesic distance on the manifold $\mathcal{M}$ rather than the linear Euclidean distance. These refined lists serve as the input for the subsequent graph construction, ensuring that the topology is corrected before any embedding learning begins.

\subsubsection*{\textbf{Stage 2: Rank-based Graph Construction and Prototype Selection}}
\label{subsubsec:stage2}

In the second stage, we model the collection as a graph $G=(V, E, \mathbf{W})$ to discover the most effective reference points (representative samples/prototypes). The graph is constructed using the refined neighborhoods from Stage 1. Unlike standard GNNs that use opaque message passing, we employ rank-based similarity measures, exemplified by RaDE \cite{fernando_rade_2020, fernando_rade_2022} and GRaCE \cite{almeida_grace_2025}, to define edge weights $\mathbf{W}_{ij}$. RaDE uses logarithmic decay ($w_{ij} = 1 - \log_k(\tau'_i(j))$), while GRaCE uses rank correlation measures (e.g., JaccardMax \cite{valem_novel_2022}).

From this graph, we select a subset of prototypes $\mathcal{R} = \{r_1, \dots, r_d\} \subset \mathcal{C}$. The selection follows a greedy optimization scheme designed to maximize \textit{Representational Coverage} while minimizing redundancy. Let $\mathcal{R}_{i-1}$ be the set of prototypes selected up to step $i-1$. The next prototype $r_i$ is chosen to maximize a marginal gain function:
\vspace{-2mm}
\begin{equation}
    r_i = \underset{c \in \mathcal{C} \setminus \mathcal{R}_{i-1}}{\arg\max} \; \frac{\eta(c)}{1 + \sum_{r \in \mathcal{R}_{i-1}} s(\tau'_c, \tau'_r)},
\end{equation}

\noindent where $\eta(c)$ estimates the node's effectiveness (e.g., via Reciprocal Affinities in RaDE and QPP measures, like Reciprocal Density \cite{pedronette_unsupervised_2015}, in GRaCE) and the denominator penalizes candidates similar to existing prototypes. This ensures the set $\mathcal{R}$ spans the diverse semantic concepts of the manifold.

\subsubsection*{\textbf{Stage 3: Interpretable Embedding Generation}}
\label{subsubsec:stage3}

The final stage generates the interpretable vector $v_e$ for every object $e$. By projecting the object onto the selected prototypes, we obtain:
\vspace{-2mm}
\begin{equation}
    v_e = [ s(\tau'_e, \tau'_{r_1}), \dots, s(\tau'_e, \tau'_{r_d}) ],
\end{equation}
\noindent where $s(\cdot)$ represents the rank-based similarity (e.g., diffusion affinity or rank correlation). 

This transformation solves the \textit{Interpretability Gap}. Unlike ``black-box'' dense layers, every dimension $i$ of $v_e$ has a transparent semantic definition: it represents the degree of similarity to the specific prototype $r_i$. Furthermore, because the input rankings $\tau'$ were refined in Stage 1, this similarity assessment respects the non-linear manifold structure, simultaneously bridging the \textit{Geometric Gap}.

\section{Experimental Setup}
\label{sec:experimental_setup}

The following subsections describe the datasets and features, evaluation protocol, and implementation details required for reproducibility.

\subsection{Datasets and Representations}
\label{subsec:datasets}

We evaluate our approach on four diverse datasets covering both general object categories and fine-grained domains: Oxford17Flowers (Flowers) \cite{nilsback_visual_2006} (17 classes, 1,360 images), Corel5k \cite{liu_content-based_2013} (50 classes, 5,000 images), Oxford-IIIT Pet (Pets) \cite{parkhi_cats-and-dogs_2012} (37 breeds, 7,390 images), and CUB-200 \cite{wah_cub200_2011} (200 species, 11,788 images).

For feature extraction, we employ three state-of-the-art architectures pre-trained on ImageNet-1K: Vision Transformer (ViT-B/16) \cite{dosovitskiy_image_2021} ($d=768$); Swin Transformer (Swin-Tf) \cite{liu_swin_2021} ($d=1024$); and ConvNeXt-Base \cite{liu2022convnet2020s} ($d=1024$). These high-dimensional vectors serve as the input for our manifold learning pipeline.

\subsection{Experimental Protocol}
\label{subsec:experimental_protocol}

We assess the effectiveness and robustness of the proposed framework across two primary tasks: Content-Based Image Retrieval (CBIR) and Semi-supervised Classification using GCNs.

\subsubsection*{Image Retrieval.} 
In the retrieval task, we evaluate the quality of the learned representations by using every image in the dataset as a query, while the remaining images constitute the retrieval gallery. The retrieval effectiveness is reported using the Mean Average Precision at depth 1000 (MAP@1000). This metric provides a comprehensive assessment of ranking quality, penalizing relevant items that appear low in the retrieved list. The depth of 1000 was adopted to align with prior works \cite{almeida_grace_2025}, ensuring comparability from the same evaluation protocol.

\subsubsection*{Semi-supervised Classification.} 
To evaluate the discriminative power of the representations in label-scarce scenarios, we employ a Graph Convolutional Network (GCN) classification protocol. Following the methodology established in \cite{valem_graph_2023}, we utilize a stratified cross-validation scheme designed to simulate semi-supervised conditions. The evaluation consists of five rounds of 10-fold cross-validation.
In each fold, only $10\%$ of the data is used as the training set, while the remaining $90\%$ serves as the test set. The classification performance is reported as the average Accuracy across these runs.

\subsection{Implementation Details and Parameters}
\label{subsec:implementation_details}

\subsubsection*{Manifold Learning.} For the manifold learning stage, we evaluated both neighbor embedding and context-aware similarity methods. For UMAP, we utilized the standard implementation with default parameters. For the Context-Aware Similarity methods, we employed three methods from the Unsupervised Distance Learning Framework (UDLF) \cite{valem2017udlf}: RDPAC \cite{pedronette_2021_rdpac}, LHRR \cite{pedronette_2019_lhrr}, and BFSTREE \cite{PEDRONETTE2021_bfstree}. For LHRR and BFSTREE, the size of the initial ranked lists was fixed at $L=1000$, and the neighborhood set size $k$ was tuned according to the specific characteristics of each dataset: $k=80$ for Oxford17Flowers, $k=100$ for Corel5k, $k=200$ for Oxford-IIIT Pet, and $k=60$ for CUB-200. Specifically, for RDPAC, the parameters were set to $L = 500$, and $k$ to the default method configuration.

\subsubsection*{Rank-based Graph Embedding.} The construction of the rank-based graph and the subsequent prototype selection relied on the RaDE \cite{fernando_rade_2020} and GRaCE \cite{almeida_grace_2025}. For graph construction, the neighborhood size parameter (denoted as $L$ in RaDE and $k$ in GRaCE) was set identically to the manifold learning stage ($k \in \{60, 80, 100, 200\}$ depending on the dataset). For RaDE, the scaling parameter was set to $t=2$. In prototype selection, we employed \textit{Reciprocal Density \cite{pedronette_unsupervised_2015}} for the node effectiveness estimation and \textit{JaccardMax \cite{valem_novel_2022}} for the rank correlation measure within GRaCE. Concerning the embedding dimensions, the target dimensionality of the interpretable embedding was set to $d=128$ for the Flowers, Corel5k, and Pets datasets, following the value adopted in the foundational works of RaDE~\cite{fernando_rade_2020, fernando_rade_2022} and GRaCE~\cite{almeida_grace_2025}, neither of which evaluated on datasets with a large number of classes. For the CUB-200 dataset (200 classes), the dimension was increased to $d=256$ to better accommodate the higher semantic diversity of this fine-grained benchmark.

\subsubsection*{GCN Configuration.} For the semi-supervised classification task, the input graph topology was constructed using a $k$-Nearest Neighbor ($k$-NN) or Reciprocal $k$-NN strategy with $k=40$. The GCN model, specifically the Simple Graph Convolution Network \cite{wu_simplifying_2019}, was optimized using the Adam optimizer for $200$ epochs. The learning rate was set to $10^{-3}$ for the Flowers, Corel5k, and Pets datasets, while for CUB-200, it was set to $10^{-2}$.

\subsubsection*{Software and Availability.} The Rank-based Graph Embedding methods were implemented using the \texttt{interpretable-embeddings} package\footnote{\href{https://github.com/thcastilho/interpretable-embeddings}{https://github.com/thcastilho/interpretable-embeddings}}. The UMAP implementation utilized the \texttt{umap-learn} library\footnote{\href{https://github.com/lmcinnes/umap}{https://github.com/lmcinnes/umap}}, and the Context-Aware Similarity Learning approaches were executed via the \texttt{pyUDLF} wrapper\footnote{\href{https://github.com/UDLF/pyUDLF}{https://github.com/UDLF/pyUDLF}}. PyTorch framework\footnote{\href{https://github.com/pyg-team/pytorch_geometric/blob/master/examples/sgc.py}{https://github.com/pyg-team/pytorch\_geometric}} was used for training the GCNs models.
\section{Results and Discussion}
\label{sec:results}

The following subsections present quantitative and qualitative analyses covering image retrieval, comparison with state-of-the-art graph embedding methods, semi-supervised classification, and interpretability.

\subsection{Retrieval}
\label{subsec:retrieval}

Results are compared against the original features and interpretable representations (RaDE, GRaCE) without manifold learning, across all dataset and backbone combinations.

Table~\ref{tab:results_flowers_final} reports the results on the \textit{Flowers} dataset. The best results for each feature are highlighted in bold, while the overall best results are shown in blue. We observe that, in addition to interpretability aspects, the retrieval effectiveness of the proposed approaches surpasses that of the original features and of interpretable representations in isolation. GRaCE, combined with rank-based manifold learning approaches, achieved the best performance.

\begin{table}[h!]
    \vspace{-2mm}
    \centering
    \caption{Retrieval results on the \textbf{Flowers} dataset, considering MAP (\%). The best result per feature is highlighted in \textbf{bold}; the best overall result is highlighted in \textbf{\color{blue}bold blue}.}
    \vspace{-3mm}
    \label{tab:results_flowers_final}
    \resizebox{0.45\textwidth}{!}{
    \begin{tabular}{llccc}
        \toprule
        \multicolumn{2}{c}{\textbf{Representation Method}} & \textbf{ConvNeXt} & \textbf{ViT-B/16} & \textbf{Swin-Tf} \\
        \midrule
        
        \multirow{3}{*}{\textbf{Baselines}} 
        & Original Features   & 92.58 & 80.06 & 92.91 \\
        & RaDE (Base)         & 98.76 & 94.95 & 99.00 \\
        & GRaCE (Base)        & 98.99 & 92.07 & 99.11 \\
        \midrule
        \multicolumn{5}{c}{\textbf{Proposed Approaches}} \\    
        \midrule
        \multirow{4}{*}{\textbf{RaDE}} 
        & + UMAP    & 98.13 & 95.75 & 98.70 \\
        & + RDPAC   & 98.34 & 95.82 & 99.53 \\
        & + LHRR    & 99.74 & 96.51 & 99.46 \\
        & + BFSTREE & 98.61 & 94.42 & 99.61 \\
        \midrule
        
        \multirow{4}{*}{\textbf{GRaCE}} 
        & + UMAP    & 98.09 & 95.77 & 98.95 \\
        & + RDPAC   & 99.28 & \textbf{97.56} & 99.81 \\
        & + LHRR    & \textbf{\color{blue}99.87} & 96.43 & 99.55 \\
        & + BFSTREE & 99.28 & 96.77 & \textbf{99.85} \\
        \bottomrule
    \end{tabular}%
    }
\vspace{4mm}
\centering
\caption{Retrieval results on the \textbf{Corel5k} dataset, considering MAP (\%). The best result per feature is highlighted in \textbf{bold}; the best overall result is highlighted in \textbf{\color{blue}bold blue}.}
\vspace{-3mm}
\label{tab:results_corel5k_final}
\resizebox{0.45\textwidth}{!}{
\begin{tabular}{llccc}
    \toprule
    \multicolumn{2}{c}{\textbf{Representation Method}} & \textbf{ConvNeXt} & \textbf{ViT-B/16} & \textbf{Swin-Tf} \\
    \midrule
    
    \multirow{3}{*}{\textbf{Baselines}} 
    & Original Features   & 80.06 & 81.47 & 73.27 \\
    & RaDE (Base)         & 88.75 & 90.55 & 90.72 \\
    & GRaCE (Base)        & 87.51 & 89.69 & 86.86 \\
    \midrule
    \multicolumn{5}{c}{\textbf{Proposed Approaches}} \\    
    \midrule    
    \multirow{4}{*}{\textbf{RaDE}} 
    & + UMAP    & 86.70 & 90.10 & 89.19 \\
    & + RDPAC   & 84.32 & 87.31 & 92.14 \\
    & + LHRR    & 94.01 & 94.10 & 96.50 \\
    & + BFSTREE & 88.09 & 91.47 & 93.33 \\
    \midrule
    
    \multirow{4}{*}{\textbf{GRaCE}} 
    & + UMAP    & 90.22 & 92.62 & 93.13 \\
    & + RDPAC   & 93.70 & 94.19 & 96.60 \\
    & + LHRR    & 94.57 & 94.23 & 96.75 \\
    & + BFSTREE & \textbf{95.08} & \textbf{95.21} & \textbf{\color{blue}96.90} \\
    \bottomrule
\end{tabular}%
}
\end{table}

Tables~\ref{tab:results_corel5k_final}, \ref{tab:results_pets_final}, and \ref{tab:results_cub200_final} present the retrieval results for the \textit{Corel5k}, \textit{Pets}, and \textit{CUB-200} datasets, respectively. The results are highly consistent across different datasets and features. Except for Swin-Tf on the Pets dataset, the best results for each feature were achieved by GRaCE combined with manifold learning methods, predominantly BFSTREE and RDPAC.
It is worth noting the substantial effectiveness gains over the original features observed in several scenarios. On the Corel5k dataset, for instance, the best performance was achieved by \textbf{GRaCE+BFSTREE}, reaching a MAP score of 96.90\%, whereas the original feature (Swin-Tf) achieved 73.27\%.

\begin{table}[h!]
\vspace{-2mm}
\centering
\caption{Retrieval results on the \textbf{Pets} dataset, considering MAP (\%). The best result per feature is highlighted in \textbf{bold}; the best overall result is highlighted in \textbf{\color{blue}bold blue}.}
\vspace{-3mm}
\label{tab:results_pets_final}
\resizebox{0.45\textwidth}{!}{
\begin{tabular}{llccc}
    \toprule
    \multicolumn{2}{c}{\textbf{Representation Method}} & \textbf{ConvNeXt} & \textbf{ViT-B/16} & \textbf{Swin-Tf} \\
    \midrule
    
    \multirow{3}{*}{\textbf{Baselines}} 
    & Original Features   & 78.59 & 83.08 & 59.06 \\
    & RaDE (Base)         & 87.04 & 89.66 & \textbf{84.70} \\
    & GRaCE (Base)        & 85.17 & 88.55 & 81.48 \\
    \midrule
     \multicolumn{5}{c}{\textbf{Proposed Approaches}} \\    
    \midrule   
    \multirow{4}{*}{\textbf{RaDE}} 
    & + UMAP    & 80.79 & 86.00 & 79.96 \\
    & + RDPAC   & 78.99 & 81.46 & 78.71 \\
    & + LHRR    & 87.21 & 89.43 & 80.36 \\
    & + BFSTREE & 82.87 & 84.58 & 79.39 \\
    \midrule
    
    \multirow{4}{*}{\textbf{GRaCE}} 
    & + UMAP    & 86.41 & \textbf{\color{blue}90.59} & 83.75 \\
    & + RDPAC   & 86.98 & 89.83 & 83.61 \\
    & + LHRR    & 87.32 & 89.67 & 80.56 \\
    & + BFSTREE & \textbf{88.04} & 89.83 & 83.59 \\
    \bottomrule
\end{tabular}%
}
\vspace{3mm}
\centering
\centering
\caption{Retrieval results on the \textbf{CUB-200} dataset, considering MAP (\%). The best result per feature is highlighted in \textbf{bold}; the best overall result is highlighted in \textbf{\color{blue}bold blue}.}
\vspace{-3mm}
\label{tab:results_cub200_final}
\resizebox{0.45\textwidth}{!}{
\begin{tabular}{llccc}
    \toprule
    \multicolumn{2}{c}{\textbf{Representation Method}} & \textbf{ConvNeXt} & \textbf{ViT-B/16} & \textbf{Swin-Tf} \\
    \midrule
    
    \multirow{3}{*}{\textbf{Baselines}} 
    & Original Features   & 61.48 & 44.95 & 58.27 \\
    & RaDE (Base)         & 65.26 & 53.03 & 70.92 \\
    & GRaCE (Base)        & 63.80 & 49.62 & 67.96 \\
    \midrule
    \multicolumn{5}{c}{\textbf{Proposed Approaches}} \\    
    \midrule    
    \multirow{4}{*}{\textbf{RaDE}} 
    & + UMAP    & 59.36 & 48.03 & 63.90 \\
    & + RDPAC   & 56.31 & 49.03 & 60.99 \\
    & + LHRR    & 66.44 & 56.91 & 70.09 \\
    & + BFSTREE & 60.77 & 53.30 & 66.23 \\
    \midrule
    
    \multirow{4}{*}{\textbf{GRaCE}} 
    & + UMAP    & 65.56 & 52.63 & 68.98 \\
    & + RDPAC   & \textbf{69.02} & 58.03 & \textbf{\color{blue}71.79} \\
    & + LHRR    & 66.93 & 57.46 & 70.45 \\
    & + BFSTREE & 68.79 & \textbf{58.53} & 71.58 \\
    \bottomrule
\end{tabular}%
}
\vspace{-3mm}
\end{table}

Despite the overall positive trend, certain backbone--manifold combinations yield degraded performance relative to their baselines, most notably RaDE paired with RDPAC on the Pets and CUB-200 datasets. GRaCE proves more resilient to this effect, consistently benefiting from manifold refinement across all datasets.

\subsection{Comparison with State-of-the-Art}
\label{subsec:state_of_the_art}

Beyond the isolated evaluation of retrieval effectiveness, we position the proposed framework against representative graph embedding strategies. We compare our method with \textbf{DeepWalk (DW)} \cite{perozzi_deepwalk_2014}, a classical topological embedding method; \textbf{DINE} \cite{piaggesi_dine_2024}, a state-of-the-art \textit{post-hoc} interpretable method; and the rank-based interpretable models, \textbf{RaDE} \cite{fernando_rade_2020} and \textbf{GRaCE} \cite{almeida_grace_2025}.

Both DeepWalk and DINE were evaluated using the implementations provided in the official DINE repository\footnote{\href{https://github.com/simonepiaggesi/dine}{https://github.com/simonepiaggesi/dine}}. To ensure a fair comparison, the embedding dimensionality $d$ and the neighborhood size $k$ used for graph construction follow the configurations detailed in Section \ref{subsec:implementation_details}.

Table \ref{tab:graph_embedding_comparision} summarizes the comparative results using ViT-B/16 features. While classical DeepWalk generally improves over the original features, it produces opaque dimensions. Conversely, applying DINE to DeepWalk results in a huge loss of effectiveness, suggesting that its sparse decomposition destroys the fine-grained manifold structure necessary for retrieval. In sharp contrast, the proposed framework systematically elevates performance beyond all baselines, achieving the highest MAP across all datasets. This confirms that, unlike DINE, our framework successfully resolves the trade-off between transparency and accuracy.

\begin{table}[t!]
\vspace{-2mm}
\centering
\caption{Comparison of Graph Embedding approaches in retrieval effectiveness (MAP \%) across all datasets with ViT-B/16 features. The best results are highlighted in \textbf{bold}.}
\vspace{-2mm}
\label{tab:graph_embedding_comparision}
\resizebox{0.45\textwidth}{!}{%
\begin{tabular}{lcccc}
    \toprule
    \textbf{Method} & \textbf{Flowers} & \textbf{Corel5k} & \textbf{Pets} & \textbf{CUB-200} \\
    \midrule
    Original Features & 80.06 & 81.47 & 83.08 & 44.95 \\
    \midrule
    DeepWalk (DW) \cite{perozzi_deepwalk_2014} & 87.11 & 86.88 & 85.94 & 47.69 \\
    DW + DINE \cite{piaggesi_dine_2024} & 45.54 & 33.40 & 41.15 & 14.83 \\
    RaDE \cite{fernando_rade_2020} & 94.95 & 90.55 & 89.66 & 53.03 \\
    GRaCE \cite{almeida_grace_2025} & 92.07 & 89.69 & 88.55 & 49.62 \\
    \midrule
    \multicolumn{5}{c}{\textbf{Proposed Approaches (Best)}} \\
    \midrule
    RaDE + Manifold & 96.51 & 94.10 & 89.43  & 56.91 \\
    GRaCE + Manifold & \textbf{97.56} & \textbf{95.21} & \textbf{90.59} & \textbf{58.53} \\
    \bottomrule
\end{tabular}%
}
\vspace{-4mm}
\end{table}

\subsection{Classification}
\label{subsec:classification}


\begin{table*}[ht!]
\caption{Accuracy (\%) of GCN semi-supervised classification on Flowers dataset.}
\vspace{-3mm}
\label{tab:results-knn-rec-flowers}
\centering
\resizebox{0.85\textwidth}{!}{
\begin{tabular}{c|c|c|c|cc|cc|cc}
\hline
\multirow{3}{*}{Input Feature} & \multirow{3}{*}{Graph Type} & \multirow{3}{*}{Input Graph} & \multirow{3}{*}{Original} & \multicolumn{6}{c}{\textbf{Proposed Approach}} \\ \cline{5-10}
 & & & & \multicolumn{2}{c|}{Manifold: None} & \multicolumn{2}{c|}{Manifold: LHRR} & \multicolumn{2}{c}{Manifold: UMAP} \\ \cline{5-10}
 & & & & GRaCE & RaDE & GRaCE & RaDE & GRaCE & RaDE \\
\hline
\multirow{6}{*}{ConvNeXt} 
 & \multirow{3}{*}{\textbf{$k$-NN}} 
 & ConvNeXt & 96.75 $\pm$ 0.02 & 98.77 $\pm$ 0.01 & 99.19 $\pm$ 0.01 & 99.28 $\pm$ 0.05 & \textbf{99.84 $\pm$ 0.01} & 99.10 $\pm$ 0.02 & 99.03 $\pm$ 0.00 \\
 &  & Swin-Tf & 97.04 $\pm$ 0.04 & 99.26 $\pm$ 0.00 & 99.47 $\pm$ 0.01 & 98.66 $\pm$ 0.07 & \textbf{99.57 $\pm$ 0.02} & 99.38 $\pm$ 0.02 & 99.39 $\pm$ 0.00 \\
 &  & ViT-B/16 & 95.50 $\pm$ 0.02 & 95.44 $\pm$ 0.03 & 95.37 $\pm$ 0.02 & 94.72 $\pm$ 0.02 & 96.44 $\pm$ 0.08 & 96.60 $\pm$ 0.04 & \textbf{97.52 $\pm$ 0.01} \\
\cline{2-10}
 & \multirow{3}{*}{\textbf{REC}} 
 & ConvNeXt & 99.48 $\pm$ 0.02 & 99.33 $\pm$ 0.02 & 99.42 $\pm$ 0.02 & 99.77 $\pm$ 0.01 & \textcolor{blue}{\textbf{99.87 $\pm$ 0.03}} & 99.21 $\pm$ 0.01 & 99.01 $\pm$ 0.01 \\
 &  & Swin-Tf & 99.70 $\pm$ 0.01 & 99.38 $\pm$ 0.01 & 99.48 $\pm$ 0.03 & \textbf{99.77 $\pm$ 0.01} & 99.76 $\pm$ 0.01 & 99.43 $\pm$ 0.01 & 99.39 $\pm$ 0.00 \\
 &  & ViT-B/16 & 97.66 $\pm$ 0.02 & 95.97 $\pm$ 0.03 & 96.13 $\pm$ 0.03 & 97.29 $\pm$ 0.02 & \textbf{97.76 $\pm$ 0.02} & 97.11 $\pm$ 0.03 & 97.61 $\pm$ 0.01 \\
\hline
\multirow{6}{*}{Swin-Tf} 
 & \multirow{3}{*}{\textbf{$k$-NN}} 
 & ConvNeXt & 96.87 $\pm$ 0.05 & 98.87 $\pm$ 0.01 & 99.26 $\pm$ 0.01 & 98.87 $\pm$ 0.10 & \textbf{99.89 $\pm$ 0.00} & 99.03 $\pm$ 0.01 & 99.04 $\pm$ 0.00 \\
 &  & Swin-Tf & 97.06 $\pm$ 0.01 & 99.32 $\pm$ 0.01 & 99.51 $\pm$ 0.01 & 98.70 $\pm$ 0.04 & \textbf{99.58 $\pm$ 0.01} & 99.42 $\pm$ 0.01 & 99.39 $\pm$ 0.00 \\
 &  & ViT-B/16 & 95.48 $\pm$ 0.03 & 95.65 $\pm$ 0.02 & 95.60 $\pm$ 0.04 & 94.80 $\pm$ 0.03 & 96.53 $\pm$ 0.07 & 96.81 $\pm$ 0.05 & \textbf{97.68 $\pm$ 0.00} \\
\cline{2-10}
 & \multirow{3}{*}{\textbf{REC}} 
 & ConvNeXt & 99.64 $\pm$ 0.02 & 99.47 $\pm$ 0.01 & 99.48 $\pm$ 0.01 & 99.83 $\pm$ 0.01 & \textcolor{blue}{\textbf{99.92 $\pm$ 0.01}} & 99.22 $\pm$ 0.01 & 99.04 $\pm$ 0.00 \\
 &  & Swin-Tf & 99.80 $\pm$ 0.01 & 99.48 $\pm$ 0.00 & 99.50 $\pm$ 0.01 & \textbf{99.88 $\pm$ 0.03} & 99.78 $\pm$ 0.01 & 99.48 $\pm$ 0.00 & 99.40 $\pm$ 0.00 \\
 &  & ViT-B/16 & 97.79 $\pm$ 0.02 & 96.13 $\pm$ 0.02 & 96.31 $\pm$ 0.02 & 97.57 $\pm$ 0.02 & \textbf{97.83 $\pm$ 0.01} & 97.36 $\pm$ 0.01 & 97.71 $\pm$ 0.01 \\
\hline
\multirow{6}{*}{ViT-B/16} 
 & \multirow{3}{*}{\textbf{$k$-NN}} 
 & ConvNeXt & 96.31 $\pm$ 0.06 & 98.89 $\pm$ 0.00 & 99.20 $\pm$ 0.01 & 96.23 $\pm$ 0.18 & \textcolor{blue}{\textbf{99.82 $\pm$ 0.02}} & 98.30 $\pm$ 0.08 & 99.01 $\pm$ 0.01 \\
 &  & Swin-Tf & 96.22 $\pm$ 0.05 & 99.25 $\pm$ 0.02 & \textbf{99.44 $\pm$ 0.03} & 96.48 $\pm$ 0.14 & 99.42 $\pm$ 0.03 & 98.85 $\pm$ 0.06 & 99.39 $\pm$ 0.00 \\
 &  & ViT-B/16 & 94.08 $\pm$ 0.03 & 95.00 $\pm$ 0.02 & 95.00 $\pm$ 0.07 & 93.39 $\pm$ 0.15 & 96.06 $\pm$ 0.08 & 95.62 $\pm$ 0.07 & \textbf{97.11 $\pm$ 0.03} \\
\cline{2-10}
 & \multirow{3}{*}{\textbf{REC}} 
 & ConvNeXt & 98.84 $\pm$ 0.02 & 99.37 $\pm$ 0.02 & 99.39 $\pm$ 0.01 & 98.68 $\pm$ 0.06 & \textbf{99.46 $\pm$ 0.01} & 98.98 $\pm$ 0.02 & 98.98 $\pm$ 0.02 \\
 &  & Swin-Tf & 98.85 $\pm$ 0.04 & 99.11 $\pm$ 0.00 & 99.39 $\pm$ 0.01 & 98.82 $\pm$ 0.05 & 99.29 $\pm$ 0.03 & 99.19 $\pm$ 0.03 & \textbf{99.40 $\pm$ 0.00} \\
 &  & ViT-B/16 & 96.50 $\pm$ 0.03 & 95.25 $\pm$ 0.03 & 95.47 $\pm$ 0.02 & 96.54 $\pm$ 0.06 & 96.96 $\pm$ 0.04 & 96.81 $\pm$ 0.03 & \textbf{97.50 $\pm$ 0.02} \\
\hline
\end{tabular}}
\end{table*}


\begin{table*}[ht!]
\caption{Accuracy (\%) of GCN semi-supervised classification on Corel5k dataset.}
\vspace{-3mm}
\label{tab:results-knn-rec-corel5k}
\centering
\resizebox{0.85\textwidth}{!}{
\begin{tabular}{c|c|c|c|cc|cc|cc}
\hline
\multirow{3}{*}{Input Feature} & \multirow{3}{*}{Graph Type} & \multirow{3}{*}{Input Graph} & \multirow{3}{*}{Original} & \multicolumn{6}{c}{\textbf{Proposed Approach}} \\ \cline{5-10}
 & & & & \multicolumn{2}{c|}{Manifold: None} & \multicolumn{2}{c|}{Manifold: LHRR} & \multicolumn{2}{c}{Manifold: UMAP} \\ \cline{5-10}
 & & & & GRaCE & RaDE & GRaCE & RaDE & GRaCE & RaDE \\
\hline
\multirow{6}{*}{ConvNeXt} 
 & \multirow{3}{*}{\textbf{$k$-NN}} 
 & ConvNeXt & 94.42 $\pm$ 0.02 & 92.69 $\pm$ 0.01 & 92.68 $\pm$ 0.03 & 93.61 $\pm$ 0.04 & \textbf{95.37 $\pm$ 0.05} & 92.86 $\pm$ 0.07 & 94.27 $\pm$ 0.08 \\
 &  & Swin-Tf & 95.99 $\pm$ 0.03 & 92.93 $\pm$ 0.02 & 94.45 $\pm$ 0.02 & 95.26 $\pm$ 0.05 & \textbf{97.18 $\pm$ 0.02} & 94.57 $\pm$ 0.14 & 96.31 $\pm$ 0.03 \\
 &  & ViT-B/16 & 95.08 $\pm$ 0.04 & 93.74 $\pm$ 0.01 & 93.75 $\pm$ 0.01 & 93.63 $\pm$ 0.03 & \textbf{95.64 $\pm$ 0.03} & 93.37 $\pm$ 0.05 & 94.59 $\pm$ 0.02 \\
\cline{2-10}
 & \multirow{3}{*}{\textbf{REC}} 
 & ConvNeXt & 96.30 $\pm$ 0.08 & 93.41 $\pm$ 0.04 & 93.40 $\pm$ 0.09 & \textbf{96.03 $\pm$ 0.04} & 95.96 $\pm$ 0.06 & 94.17 $\pm$ 0.03 & 94.46 $\pm$ 0.03 \\
 &  & Swin-Tf & 97.39 $\pm$ 0.05 & 94.06 $\pm$ 0.02 & 95.06 $\pm$ 0.04 & 96.98 $\pm$ 0.03 & \textcolor{blue}{\textbf{97.45 $\pm$ 0.06}} & 95.95 $\pm$ 0.06 & 96.63 $\pm$ 0.06 \\
 &  & ViT-B/16 & 96.47 $\pm$ 0.01 & 94.37 $\pm$ 0.01 & 94.60 $\pm$ 0.02 & 95.30 $\pm$ 0.03 & \textbf{96.10 $\pm$ 0.09} & 95.17 $\pm$ 0.05 & 95.02 $\pm$ 0.03 \\
\hline
\multirow{6}{*}{Swin-Tf} 
 & \multirow{3}{*}{\textbf{$k$-NN}} 
 & ConvNeXt & 94.62 $\pm$ 0.03 & 92.89 $\pm$ 0.01 & 93.37 $\pm$ 0.02 & 93.75 $\pm$ 0.05 & \textbf{95.67 $\pm$ 0.02} & 93.35 $\pm$ 0.05 & 94.77 $\pm$ 0.06 \\
 &  & Swin-Tf & 95.75 $\pm$ 0.04 & 93.12 $\pm$ 0.02 & 94.47 $\pm$ 0.05 & 95.81 $\pm$ 0.05 & \textbf{97.22 $\pm$ 0.01} & 95.14 $\pm$ 0.04 & 96.46 $\pm$ 0.02 \\
 &  & ViT-B/16 & 95.33 $\pm$ 0.03 & 93.87 $\pm$ 0.01 & 94.06 $\pm$ 0.04 & 93.69 $\pm$ 0.05 & \textbf{95.83 $\pm$ 0.02} & 93.71 $\pm$ 0.06 & 94.77 $\pm$ 0.02 \\
\cline{2-10}
 & \multirow{3}{*}{\textbf{REC}} 
 & ConvNeXt & 97.09 $\pm$ 0.05 & 93.88 $\pm$ 0.02 & 94.27 $\pm$ 0.07 & \textbf{96.74 $\pm$ 0.05} & 96.49 $\pm$ 0.03 & 94.79 $\pm$ 0.05 & 95.01 $\pm$ 0.02 \\
 &  & Swin-Tf & 97.76 $\pm$ 0.04 & 94.20 $\pm$ 0.01 & 95.18 $\pm$ 0.11 & 97.56 $\pm$ 0.02 & \textcolor{blue}{\textbf{97.65 $\pm$ 0.04}} & 96.43 $\pm$ 0.03 & 96.79 $\pm$ 0.01 \\
 &  & ViT-B/16 & 96.84 $\pm$ 0.01 & 94.69 $\pm$ 0.01 & 94.88 $\pm$ 0.03 & 96.02 $\pm$ 0.02 & \textbf{96.41 $\pm$ 0.06} & 95.78 $\pm$ 0.01 & 95.20 $\pm$ 0.03 \\
\hline
\multirow{6}{*}{ViT-B/16} 
 & \multirow{3}{*}{\textbf{$k$-NN}} 
 & ConvNeXt & 94.56 $\pm$ 0.04 & 92.75 $\pm$ 0.01 & 92.87 $\pm$ 0.05 & 93.44 $\pm$ 0.07 & \textbf{95.44 $\pm$ 0.01} & 92.95 $\pm$ 0.07 & 94.40 $\pm$ 0.12 \\
 &  & Swin-Tf & 95.81 $\pm$ 0.03 & 92.93 $\pm$ 0.02 & 94.48 $\pm$ 0.06 & 95.23 $\pm$ 0.06 & \textbf{97.16 $\pm$ 0.02} & 94.39 $\pm$ 0.03 & 96.43 $\pm$ 0.03 \\
 &  & ViT-B/16 & 94.64 $\pm$ 0.04 & 93.60 $\pm$ 0.01 & 93.65 $\pm$ 0.03 & 93.23 $\pm$ 0.10 & \textbf{95.65 $\pm$ 0.02} & 93.29 $\pm$ 0.08 & 94.48 $\pm$ 0.05 \\
\cline{2-10}
 & \multirow{3}{*}{\textbf{REC}} 
 & ConvNeXt & 96.61 $\pm$ 0.09 & 93.57 $\pm$ 0.02 & 93.72 $\pm$ 0.17 & 95.93 $\pm$ 0.02 & \textbf{96.10 $\pm$ 0.06} & 94.41 $\pm$ 0.06 & 94.56 $\pm$ 0.09 \\
 &  & Swin-Tf & 97.35 $\pm$ 0.05 & 94.01 $\pm$ 0.03 & 95.10 $\pm$ 0.06 & 96.72 $\pm$ 0.02 & \textcolor{blue}{\textbf{97.31 $\pm$ 0.10}} & 95.71 $\pm$ 0.07 & 96.62 $\pm$ 0.06 \\
 &  & ViT-B/16 & 96.24 $\pm$ 0.02 & 94.19 $\pm$ 0.02 & 94.30 $\pm$ 0.10 & 94.81 $\pm$ 0.03 & \textbf{95.81 $\pm$ 0.06} & 94.99 $\pm$ 0.05 & 94.87 $\pm$ 0.07 \\
\hline
\end{tabular}}
\end{table*}

The proposed context-aware representations were also evaluated for computing $k$-NN and Reciprocal $k$-NN (REC) graphs exploited for semi-supervised classification using GCNs~\cite{kipf_semi-supervised_2017}. The experiments compared graphs computed from the original high-dimensional features with those derived from the interpretable embeddings, evaluating the impact of the manifold learning step. The best results obtained with interpretable representations are highlighted in bold within each row and in blue for each feature.

Table~\ref{tab:results-knn-rec-flowers} presents the results on the \textit{Flowers} dataset. The highest accuracy scores were achieved by RaDE interpretable representations combined with rank-based manifold learning using LHRR. Table~\ref{tab:results-knn-rec-corel5k} reports the results for the \textit{Corel5k} dataset, where similar trends were observed.
Tables~\ref{tab:results-knn-rec-pets} and \ref{tab:results-knn-rec-cub200} present the classification results for the \textit{Pets} and \textit{CUB-200} datasets, respectively. In these fine-grained scenarios, RaDE representations also achieved the best performance, predominantly when combined with LHRR and UMAP manifold learning strategies.

Overall, when comparing the interpretable representations with the original features, we observe that classification accuracy is robustly maintained in most scenarios. This indicates that the proposed approach effectively compresses the feature space (e.g., from 1024 to 128 dimensions) and provides semantic interpretability without compromising the discriminative power required for downstream classification tasks.

\begin{table*}[ht!]
\caption{Accuracy (\%) of GCN semi-supervised classification on Pets dataset.}
\vspace{-3mm}
\label{tab:results-knn-rec-pets}
\centering
\resizebox{0.85\textwidth}{!}{
\begin{tabular}{c|c|c|c|cc|cc|cc}
\hline
\multirow{3}{*}{Input Feature} & \multirow{3}{*}{Graph Type} & \multirow{3}{*}{Input Graph} & \multirow{3}{*}{Original} & \multicolumn{6}{c}{\textbf{Proposed Approach}} \\ \cline{5-10}
 & & & & \multicolumn{2}{c|}{Manifold: None} & \multicolumn{2}{c|}{Manifold: LHRR} & \multicolumn{2}{c}{Manifold: UMAP} \\ \cline{5-10}
 & & & & GRaCE & RaDE & GRaCE & RaDE & GRaCE & RaDE \\
\hline
\multirow{6}{*}{ConvNeXt} 
 & \multirow{3}{*}{\textbf{$k$-NN}} 
 & ConvNeXt & 88.52 $\pm$ 0.03 & 91.46 $\pm$ 0.01 & \textbf{92.12 $\pm$ 0.02} & 91.13 $\pm$ 0.02 & 91.59 $\pm$ 0.01 & 90.35 $\pm$ 0.04 & 90.73 $\pm$ 0.04 \\
 &  & Swin-Tf & 85.90 $\pm$ 0.06 & 88.69 $\pm$ 0.02 & \textbf{89.82 $\pm$ 0.02} & 85.62 $\pm$ 0.02 & 85.43 $\pm$ 0.02 & 87.88 $\pm$ 0.07 & 87.75 $\pm$ 0.02 \\
 &  & ViT-B/16 & 91.16 $\pm$ 0.01 & 92.57 $\pm$ 0.01 & 92.56 $\pm$ 0.02 & 91.33 $\pm$ 0.03 & 92.63 $\pm$ 0.01 & 93.19 $\pm$ 0.04 & \textbf{93.34 $\pm$ 0.02} \\
\cline{2-10}
 & \multirow{3}{*}{\textbf{REC}} 
 & ConvNeXt & 92.14 $\pm$ 0.01 & 91.81 $\pm$ 0.01 & \textbf{92.44 $\pm$ 0.01} & 91.66 $\pm$ 0.02 & 91.69 $\pm$ 0.02 & 91.10 $\pm$ 0.03 & 91.12 $\pm$ 0.04 \\
 &  & Swin-Tf & 91.02 $\pm$ 0.03 & \textbf{89.05 $\pm$ 0.01} & 90.05 $\pm$ 0.02 & 86.65 $\pm$ 0.02 & 86.87 $\pm$ 0.01 & 88.16 $\pm$ 0.02 & 88.18 $\pm$ 0.02 \\
 &  & ViT-B/16 & 93.27 $\pm$ 0.03 & 92.72 $\pm$ 0.01 & 92.96 $\pm$ 0.01 & 91.90 $\pm$ 0.01 & 92.95 $\pm$ 0.02 & 93.34 $\pm$ 0.01 & \textcolor{blue}{\textbf{93.50 $\pm$ 0.02}} \\
\hline
\multirow{6}{*}{Swin-Tf} 
 & \multirow{3}{*}{\textbf{$k$-NN}} 
 & ConvNeXt & 87.84 $\pm$ 0.05 & 91.64 $\pm$ 0.01 & \textbf{92.21 $\pm$ 0.01} & 90.63 $\pm$ 0.03 & 91.67 $\pm$ 0.01 & 89.78 $\pm$ 0.01 & 90.85 $\pm$ 0.03 \\
 &  & Swin-Tf & 83.92 $\pm$ 0.04 & 88.62 $\pm$ 0.00 & \textbf{89.64 $\pm$ 0.02} & 85.65 $\pm$ 0.02 & 85.39 $\pm$ 0.02 & 87.57 $\pm$ 0.06 & 87.74 $\pm$ 0.01 \\
 &  & ViT-B/16 & 90.90 $\pm$ 0.02 & 92.65 $\pm$ 0.00 & 92.61 $\pm$ 0.01 & 90.65 $\pm$ 0.03 & 92.68 $\pm$ 0.01 & 92.41 $\pm$ 0.03 & \textbf{93.22 $\pm$ 0.02} \\
\cline{2-10}
 & \multirow{3}{*}{\textbf{REC}} 
 & ConvNeXt & 91.76 $\pm$ 0.03 & 91.88 $\pm$ 0.02 & \textbf{92.57 $\pm$ 0.02} & 91.48 $\pm$ 0.03 & 92.09 $\pm$ 0.02 & 91.06 $\pm$ 0.03 & 91.34 $\pm$ 0.02 \\
 &  & Swin-Tf & 89.61 $\pm$ 0.04 & 88.80 $\pm$ 0.01 & \textbf{89.79 $\pm$ 0.00} & 86.45 $\pm$ 0.01 & 86.60 $\pm$ 0.01 & 87.89 $\pm$ 0.03 & 88.11 $\pm$ 0.00 \\
 &  & ViT-B/16 & 92.91 $\pm$ 0.02 & 92.84 $\pm$ 0.01 & 93.01 $\pm$ 0.01 & 91.44 $\pm$ 0.02 & 92.92 $\pm$ 0.00 & 92.89 $\pm$ 0.02 & \textcolor{blue}{\textbf{93.33 $\pm$ 0.03}} \\
\hline
\multirow{6}{*}{ViT-B/16} 
 & \multirow{3}{*}{\textbf{$k$-NN}} 
 & ConvNeXt & 88.81 $\pm$ 0.04 & 91.52 $\pm$ 0.01 & \textbf{92.20 $\pm$ 0.01} & 91.06 $\pm$ 0.05 & 91.51 $\pm$ 0.03 & 90.17 $\pm$ 0.04 & 90.51 $\pm$ 0.02 \\
 &  & Swin-Tf & 86.08 $\pm$ 0.04 & 88.56 $\pm$ 0.02 & \textbf{89.81 $\pm$ 0.02} & 85.58 $\pm$ 0.02 & 85.22 $\pm$ 0.02 & 87.56 $\pm$ 0.08 & 87.45 $\pm$ 0.04 \\
 &  & ViT-B/16 & 91.09 $\pm$ 0.04 & 92.44 $\pm$ 0.01 & 92.25 $\pm$ 0.06 & 91.47 $\pm$ 0.04 & 92.50 $\pm$ 0.01 & \textbf{93.10 $\pm$ 0.04} & 93.02 $\pm$ 0.02 \\
\cline{2-10}
 & \multirow{3}{*}{\textbf{REC}} 
 & ConvNeXt & 92.47 $\pm$ 0.02 & 91.92 $\pm$ 0.02 & \textbf{92.45 $\pm$ 0.01} & 91.66 $\pm$ 0.04 & 91.92 $\pm$ 0.02 & 91.03 $\pm$ 0.04 & 90.87 $\pm$ 0.04 \\
 &  & Swin-Tf & 91.08 $\pm$ 0.05 & 89.00 $\pm$ 0.01 & \textbf{90.09 $\pm$ 0.01} & 86.58 $\pm$ 0.02 & 86.76 $\pm$ 0.02 & 87.78 $\pm$ 0.02 & 87.90 $\pm$ 0.06 \\
 &  & ViT-B/16 & 92.87 $\pm$ 0.03 & 92.57 $\pm$ 0.01 & 92.58 $\pm$ 0.04 & 91.84 $\pm$ 0.01 & 92.60 $\pm$ 0.03 & \textcolor{blue}{\textbf{93.15 $\pm$ 0.02}} & 93.09 $\pm$ 0.08 \\
\hline
\end{tabular}}
\end{table*}


\begin{table*}[ht!]
\caption{Accuracy (\%) of GCN semi-supervised classification on CUB-200 dataset.}
\vspace{-3mm}
\label{tab:results-knn-rec-cub200}
\centering
\resizebox{0.85\textwidth}{!}{
\begin{tabular}{c|c|c|c|cc|cc|cc}
\hline
\multirow{3}{*}{Input Feature} & \multirow{3}{*}{Graph Type} & \multirow{3}{*}{Input Graph} & \multirow{3}{*}{Original} & \multicolumn{6}{c}{\textbf{Proposed Approach}} \\ \cline{5-10}
 & & & & \multicolumn{2}{c|}{Manifold: None} & \multicolumn{2}{c|}{Manifold: LHRR} & \multicolumn{2}{c}{Manifold: UMAP} \\ \cline{5-10}
 & & & & GRaCE & RaDE & GRaCE & RaDE & GRaCE & RaDE \\
\hline
\multirow{6}{*}{ConvNeXt} 
 & \multirow{3}{*}{\textbf{$k$-NN}} 
 & ConvNeXt & 76.96 $\pm$ 0.03 & 73.24 $\pm$ 0.03 & 74.85 $\pm$ 0.02 & 75.84 $\pm$ 0.06 & \textbf{75.98 $\pm$ 0.04} & 75.07 $\pm$ 0.03 & 73.99 $\pm$ 0.03 \\
 &  & Swin-Tf & 77.90 $\pm$ 0.02 & 75.97 $\pm$ 0.02 & 77.91 $\pm$ 0.03 & 77.37 $\pm$ 0.04 & 77.41 $\pm$ 0.02 & \textbf{77.98 $\pm$ 0.09} & 77.26 $\pm$ 0.03 \\
 &  & ViT-B/16 & 71.33 $\pm$ 0.03 & 63.32 $\pm$ 0.01 & 66.60 $\pm$ 0.02 & 68.51 $\pm$ 0.05 & \textbf{68.78 $\pm$ 0.03} & 65.25 $\pm$ 0.04 & 64.55 $\pm$ 0.03 \\
\cline{2-10}
 & \multirow{3}{*}{\textbf{REC}} 
 & ConvNeXt & 80.29 $\pm$ 0.02 & 73.64 $\pm$ 0.04 & 75.31 $\pm$ 0.02 & 77.08 $\pm$ 0.09 & \textbf{77.51 $\pm$ 0.05} & 75.99 $\pm$ 0.05 & 75.35 $\pm$ 0.06 \\
 &  & Swin-Tf & 82.17 $\pm$ 0.02 & 76.95 $\pm$ 0.02 & 78.72 $\pm$ 0.04 & 78.86 $\pm$ 0.03 & \textcolor{blue}{\textbf{79.39 $\pm$ 0.03}} & 78.77 $\pm$ 0.02 & 78.50 $\pm$ 0.02 \\
 &  & ViT-B/16 & 76.27 $\pm$ 0.01 & 64.83 $\pm$ 0.01 & 67.67 $\pm$ 0.03 & 70.63 $\pm$ 0.03 & \textbf{70.91 $\pm$ 0.04} & 66.52 $\pm$ 0.02 & 65.64 $\pm$ 0.02 \\
\hline
\multirow{6}{*}{Swin-Tf} 
 & \multirow{3}{*}{\textbf{$k$-NN}} 
 & ConvNeXt & 77.67 $\pm$ 0.00 & 73.72 $\pm$ 0.02 & 75.52 $\pm$ 0.03 & 76.21 $\pm$ 0.03 & \textbf{76.52 $\pm$ 0.05} & 75.31 $\pm$ 0.04 & 74.55 $\pm$ 0.02 \\
 &  & Swin-Tf & 77.54 $\pm$ 0.01 & 76.24 $\pm$ 0.03 & \textbf{78.18 $\pm$ 0.02} & 77.38 $\pm$ 0.04 & 77.80 $\pm$ 0.01 & 77.94 $\pm$ 0.04 & 77.54 $\pm$ 0.03 \\
 &  & ViT-B/16 & 71.91 $\pm$ 0.01 & 63.87 $\pm$ 0.02 & 67.02 $\pm$ 0.03 & 68.92 $\pm$ 0.02 & \textbf{69.13 $\pm$ 0.03} & 65.75 $\pm$ 0.03 & 65.00 $\pm$ 0.02 \\
\cline{2-10}
 & \multirow{3}{*}{\textbf{REC}} 
 & ConvNeXt & 81.89 $\pm$ 0.01 & 74.59 $\pm$ 0.02 & 76.08 $\pm$ 0.01 & \textbf{77.98 $\pm$ 0.02} & 78.45 $\pm$ 0.03 & 76.65 $\pm$ 0.03 & 75.91 $\pm$ 0.02 \\
 &  & Swin-Tf & 82.05 $\pm$ 0.01 & 77.16 $\pm$ 0.02 & 78.85 $\pm$ 0.03 & \textcolor{blue}{\textbf{79.20 $\pm$ 0.01}} & 79.63 $\pm$ 0.01 & 79.19 $\pm$ 0.02 & 78.75 $\pm$ 0.03 \\
 &  & ViT-B/16 & 76.75 $\pm$ 0.0 & 65.58 $\pm$ 0.01 & 68.26 $\pm$ 0.02 & \textbf{71.11 $\pm$ 0.03} & 71.46 $\pm$ 0.04 & 67.15 $\pm$ 0.03 & 66.03 $\pm$ 0.01 \\
\hline
\multirow{6}{*}{ViT-B/16} 
 & \multirow{3}{*}{\textbf{$k$-NN}} 
 & ConvNeXt & 76.89 $\pm$ 0.04 & 72.90 $\pm$ 0.02 & 74.62 $\pm$ 0.02 & 75.27 $\pm$ 0.04 & \textbf{75.81 $\pm$ 0.07} & 74.18 $\pm$ 0.04 & 73.70 $\pm$ 0.01 \\
 &  & Swin-Tf & 76.79 $\pm$ 0.07 & 75.46 $\pm$ 0.02 & \textbf{77.51 $\pm$ 0.01} & 76.74 $\pm$ 0.06 & 77.05 $\pm$ 0.04 & 77.04 $\pm$ 0.06 & 76.61 $\pm$ 0.02 \\
 &  & ViT-B/16 & 69.33 $\pm$ 0.03 & 62.49 $\pm$ 0.02 & 65.64 $\pm$ 0.01 & 67.69 $\pm$ 0.04 & \textbf{68.10 $\pm$ 0.03} & 64.28 $\pm$ 0.03 & 63.78 $\pm$ 0.02 \\
\cline{2-10}
 & \multirow{3}{*}{\textbf{REC}} 
 & ConvNeXt & 80.01 $\pm$ 0.02 & 73.16 $\pm$ 0.02 & 75.03 $\pm$ 0.02 & \textbf{75.90 $\pm$ 0.02} & 77.07 $\pm$ 0.02 & 74.85 $\pm$ 0.03 & 74.85 $\pm$ 0.01 \\
 &  & Swin-Tf & 80.15 $\pm$ 0.02 & 76.11 $\pm$ 0.01 & 78.15 $\pm$ 0.02 & 77.58 $\pm$ 0.03 & \textcolor{blue}{\textbf{78.44 $\pm$ 0.02}} & 77.54 $\pm$ 0.05 & 77.44 $\pm$ 0.03 \\
 &  & ViT-B/16 & 72.55 $\pm$ 0.02 & 63.39 $\pm$ 0.00 & 66.34 $\pm$ 0.01 & 68.87 $\pm$ 0.02 & \textbf{69.45 $\pm$ 0.04} & 64.94 $\pm$ 0.03 & 64.16 $\pm$ 0.02 \\
\hline
\end{tabular}}
\vspace{-3mm}
\end{table*}

\subsection{Qualitative and Visual Analysis}
\label{subsec:visual_results}

\begin{figure}[H]
    \centering
    \includegraphics[width=1\linewidth]{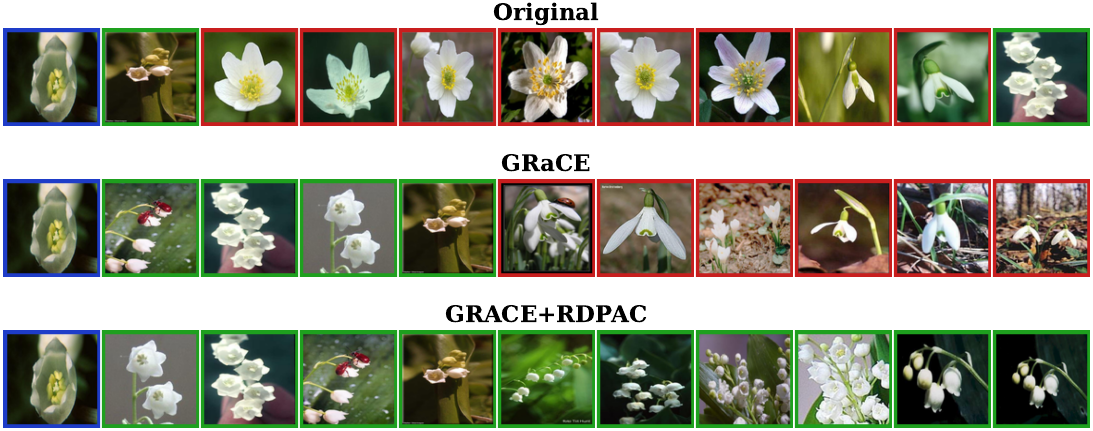}
    \vspace{-4mm}
    \caption{Qualitative retrieval results.}
    \label{fig:flowers_visual_rks}
    \vspace{-3mm}
\end{figure}

In addition to the quantitative evaluation focused on the effectiveness of downstream tasks, the experimental analysis also considered qualitative aspects. 
Figure~\ref{fig:flowers_visual_rks} presents a visual analysis comparing the retrieval results obtained using the original features, the interpretable techniques, and the proposed approach, which combines manifold learning and interpretable representations. The reported examples refer to ViT/B16 features on the \textit{Flowers} dataset, considering GRaCE in isolation and GRaCE+RDPAC for the proposed approach. The blue border indicates the query image, green borders denote relevant images, and red borders indicate non-relevant ones. Significant effectiveness gains can be observed.

\begin{figure*}[ht!]
    \centering
    \includegraphics[width=0.95\linewidth]{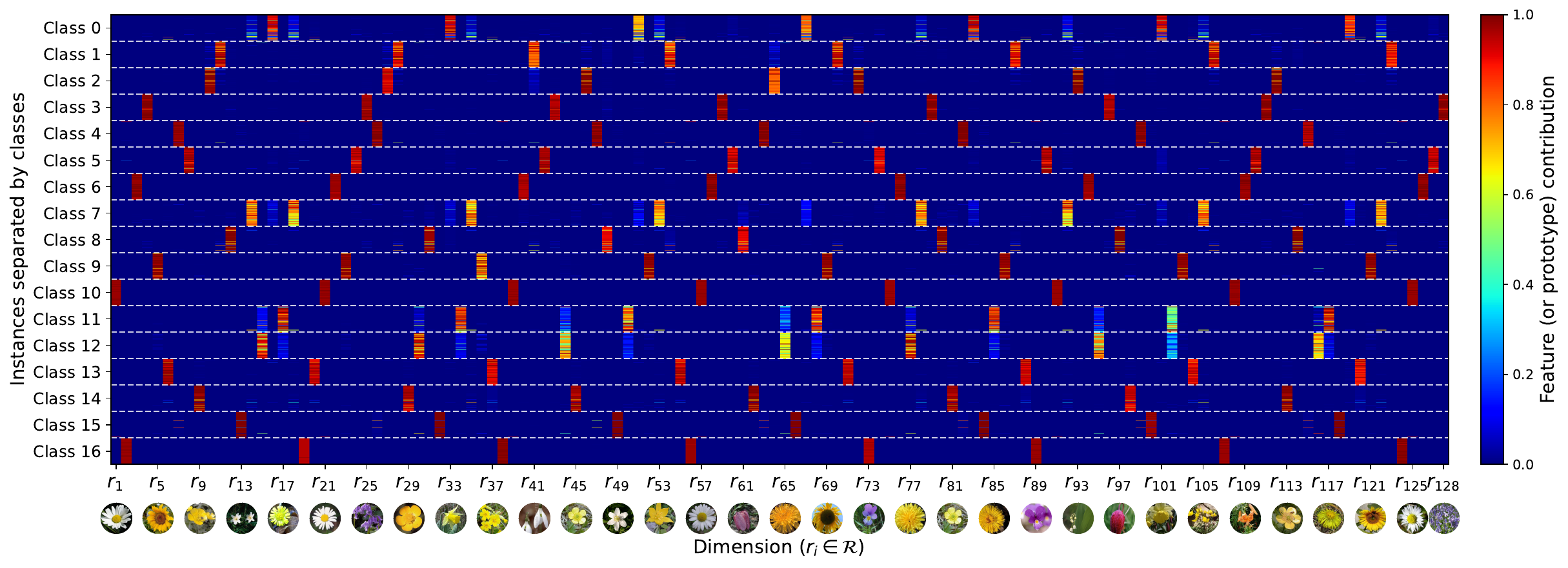}
    \vspace{-5mm}
    \caption{Heatmap visualization of the interpretable representation created by the GRaCE+RDPAC combination of the proposed approach. Each row corresponds to a dataset instance, and columns represent dimensions associated with prototypes. The color scale indicates the similarity between each instance and the corresponding prototype.}
    \label{fig:heatmap}
    \vspace{-3mm}
\end{figure*}

Figure~\ref{fig:heatmap} presents a visualization of the interpretable representation. Each row illustrates the representation obtained for a dataset instance, organized according to dataset classes (species). The columns represent the dimensions associated with prototypes. At the bottom of the figure, selected prototypes are illustrated. The color scale represents low or high values assigned to each dimension and, therefore, the similarity to the corresponding prototype.
The association between dimensions and prototypes enables a clear, human-guided interpretation of the representations. In addition, strong evidence of disentanglement can be observed. The matrix is predominantly inactive (blue), indicating quantitative sparsity and the isolation of semantic concepts associated with specific dimensions.

\subsection{Limitations}
\label{subsec:limitations}

While the proposed framework demonstrates consistent gains, some limitations should be acknowledged. First, the approach exhibits limited scalability in dynamic scenarios: incorporating new samples requires recomputing ranked lists, a cost that grows linearly with dataset size. Second, the neighborhood parameters $k$ and $L$ are adjusted per dataset and lack an automatic selection strategy, requiring practitioner tuning. Third, certain backbone--manifold combinations yield degraded performance, suggesting sensitivity to the geometric structure of specific feature spaces that warrants further investigation.

\section{Conclusions}
\label{sec:conclusion}

This paper introduced a novel framework for context-aware interpretable representations, effectively combining rank-based manifold learning with prototype-driven embedding generation. Our core contribution lies in bridging the geometric and interpretability gaps, transforming high-dimensional, opaque feature spaces into sparse, self-explaining structures that remain faithful to the intrinsic data topology.
Experimental results across diverse benchmarks validate that the proposed approach either preserves or improves effectiveness in image retrieval and semi-supervised classification via Graph Convolutional Networks (GCNs). Beyond quantitative gains, the qualitative-visual analysis confirms that the identified prototypes provide intuitive, human-centric explanations.

Future research will investigate the limitations while evaluating the integration of self-supervised contrastive learning strategies upon the interpretable representations, aiming to further distill discriminative features while maintaining the structural transparency of the embeddings.

\begin{acks}
The authors are grateful to the National Council for Scientific and Technological Development --- CNPq (grant \#313193/2023-1), the São Paulo Research Foundation --- FAPESP (grant \#2024/04890-5 and \#2025/07171-2), and Petrobras (grant \#2023/00095-3) for their financial support.
\end{acks}

\bibliographystyle{ACM-Reference-Format}
\bibliography{references}

@INPROCEEDINGS{almeida_grace_2025,
  author={Almeida, Thiago César Castilho and Rosseto Letício, Gustavo and Valem, Lucas Pascotti and Freitas, André and Guimarães Pedronette, Daniel Carlos},
  booktitle={2025 International Joint Conference on Neural Networks (IJCNN)}, 
  title={Effective Graph and Rank-based Contextual Embeddings for Textual and Multimedia Data}, 
  year={2025},
  volume={},
  number={},
  pages={1-8},
  doi={10.1109/IJCNN64981.2025.11229362}}

@conference{fernando_rade_2020,
    author={Filipe {Alves de Fernando} and Daniel Carlos Guimarães Pedronette and Gustavo {José de Sousa} and Lucas Pascotti Valem and Ivan Rizzo Guilherme},
    title={RaDE: A Rank-based Graph Embedding Approach},
    booktitle={Proceedings of the 15th International Joint Conference on Computer Vision, Imaging and Computer Graphics Theory and Applications (VISIGRAPP 2020) - Volume 5: VISAPP},
    year={2020},
    pages={142-152},
    publisher={SciTePress},
    organization={INSTICC},
    doi={10.5220/0008985901420152},
}

@inproceedings{anand2022explainable,
author = {Anand, Avishek and Saha, Sourav and Venktesh, V.},
title = {Explainable Information Retrieval},
year = {2025},
isbn = {978-3-031-88719-2},
publisher = {Springer-Verlag},
address = {Berlin, Heidelberg},
booktitle = {Advances in Information Retrieval: 47th European Conference on Information Retrieval, ECIR 2025},
pages = {254–261},
numpages = {8},
location = {Lucca, Italy}
}

@article{ANGELOV2025IDEAL,
title = {IDEAL: Interpretable-by-Design ALgorithms for learning from foundation feature spaces},
journal = {Neurocomputing},
volume = {626},
pages = {129464},
year = {2025},
doi = {10.1016/j.neucom.2025.129464},
author = {Plamen Angelov and Dmitry Kangin and Ziyang Zhang},
}

@ARTICLE{bengio2014representationlearningreviewnew,
  author={Bengio, Yoshua and Courville, Aaron and Vincent, Pascal},
  journal={IEEE Transactions on Pattern Analysis and Machine Intelligence}, 
  title={Representation Learning: A Review and New Perspectives}, 
  year={2013},
  volume={35},
  number={8},
  pages={1798-1828},
  doi={10.1109/TPAMI.2013.50}}

@inproceedings{chen2019protopnet,
 author = {Chen, Chaofan and Li, Oscar and Tao, Daniel and Barnett, Alina and Rudin, Cynthia and Su, Jonathan K},
 booktitle = {Advances in Neural Information Processing Systems},
 pages = {},
 publisher = {Curran Associates, Inc.},
 title = {This Looks Like That: Deep Learning for Interpretable Image Recognition},
 url = {https://proceedings.neurips.cc/paper_files/paper/2019/file/adf7ee2dcf142b0e11888e72b43fcb75-Paper.pdf},
 volume = {32},
 year = {2019}
}

@article{Chen_2020_graphrepresentationlearningasurvey,
   title={Graph representation learning: a survey},
   volume={9},
   url={http://dx.doi.org/10.1017/ATSIP.2020.13},
   DOI={10.1017/atsip.2020.13},
   number={1},
   journal={APSIPA Transactions on Signal and Information Processing},
   publisher={Now Publishers},
   author={Chen, Fenxiao and Wang, Yun-Cheng and Wang, Bin and Kuo, C.-C. Jay},
   year={2020} }

@inproceedings{
dosovitskiy_image_2021,
title={An Image is Worth 16x16 Words: Transformers for Image Recognition at Scale},
author={Alexey Dosovitskiy and Lucas Beyer and Alexander Kolesnikov and Dirk Weissenborn and Xiaohua Zhai and Thomas Unterthiner and Mostafa Dehghani and Matthias Minderer and Georg Heigold and Sylvain Gelly and Jakob Uszkoreit and Neil Houlsby},
booktitle={International Conference on Learning Representations},
year={2021},
}

@INPROCEEDINGS{ermolov_hyperbolic_2022,
  author={Ermolov, Aleksandr and Mirvakhabova, Leyla and Khrulkov, Valentin and Sebe, Nicu and Oseledets, Ivan},
  booktitle={2022 IEEE/CVF Conference on Computer Vision and Pattern Recognition (CVPR)}, 
  title={Hyperbolic Vision Transformers: Combining Improvements in Metric Learning}, 
  year={2022},
  volume={},
  number={},
  pages={7399-7409},
  doi={10.1109/CVPR52688.2022.00726}}

@article{fernando_rade_2022,
	title = {{RaDE}+: {A} semantic rank-based graph embedding algorithm},
	volume = {2},
	doi = {10.1016/j.jjimei.2022.100078},
	number = {1},
	journal = {International Journal of Information Management Data Insights},
	author = {Fernando, Filipe Alves de and Pedronette, Daniel Carlos Guimarães and Sousa, Gustavo José de and Valem, Lucas Pascotti and Guilherme, Ivan Rizzo},
	year = {2022},
	pages = {100078},
}

@article{
gautam2024prototypical,
title={Prototypical Self-Explainable Models Without Re-training},
author={Srishti Gautam and Ahcene Boubekki and Marina MC H{\"o}hne and Michael Kampffmeyer},
journal={Transactions on Machine Learning Research},
issn={2835-8856},
year={2024},
}

@inproceedings{grover_node2vec_2016,
	series = {{KDD} '16},
	title = {node2vec: {Scalable} {Feature} {Learning} for {Networks}},
	doi = {10.1145/2939672.2939754},
	booktitle = {Proceedings of the 22nd {ACM} {SIGKDD} {International} {Conference} on {Knowledge} {Discovery} and {Data} {Mining}},
	publisher = {ACM},
	author = {Grover, Aditya and Leskovec, Jure},
	year = {2016},
	pages = {855--864},
}

@inproceedings{he_deep_2016,
	title = {Deep {Residual} {Learning} for {Image} {Recognition}},
	booktitle = {2016 {IEEE} {Conference} on {Computer} {Vision} and {Pattern} {Recognition} ({CVPR})},
	author = {He, Kaiming and Zhang, Xiangyu and Ren, Shaoqing and Sun, Jian},
	year = {2016},
	pages = {770--778},
doi = {10.1109/CVPR.2016.90}
}

@inproceedings{
higgins2017beta,
title={beta-{VAE}: Learning Basic Visual Concepts with a Constrained Variational Framework},
author={Irina Higgins and Loic Matthey and Arka Pal and Christopher Burgess and Xavier Glorot and Matthew Botvinick and Shakir Mohamed and Alexander Lerchner},
booktitle={International Conference on Learning Representations},
year={2017},
url={https://openreview.net/forum?id=Sy2fzU9gl}
}

@inproceedings{hou2022graphmaeselfsupervisedmaskedgraph,
author = {Hou, Zhenyu and Liu, Xiao and Cen, Yukuo and Dong, Yuxiao and Yang, Hongxia and Wang, Chunjie and Tang, Jie},
title = {GraphMAE: Self-Supervised Masked Graph Autoencoders},
year = {2022},
isbn = {9781450393850},
publisher = {Association for Computing Machinery},
address = {New York, NY, USA},
url = {https://doi.org/10.1145/3534678.3539321},
doi = {10.1145/3534678.3539321},
booktitle = {Proceedings of the 28th ACM SIGKDD Conference on Knowledge Discovery and Data Mining},
pages = {594–604},
numpages = {11},
location = {Washington DC, USA},
series = {KDD '22}
}

@article{huang_review_2019,
	title = {A {Review} on {Dimensionality} {Reduction} {Techniques}},
	volume = {33},
	doi = {10.1142/S0218001419500174},
	number = {10},
	journal = {International Journal of Pattern Recognition and Artificial Intelligence},
	author = {Huang, Xuan and Wu, Lei and Ye, Yinsong},
	year = {2019},
	pages = {1950017},
}

@inproceedings{
kipf_semi-supervised_2017,
title={Semi-Supervised Classification with Graph Convolutional Networks},
author={Thomas N. Kipf and Max Welling},
booktitle={International Conference on Learning Representations},
year={2017},
url={https://openreview.net/forum?id=SJU4ayYgl}
}

@InProceedings{koh2020conceptbottleneckmodels,
  title = 	 {Concept Bottleneck Models},
  author =       {Koh, Pang Wei and Nguyen, Thao and Tang, Yew Siang and Mussmann, Stephen and Pierson, Emma and Kim, Been and Liang, Percy},
  booktitle = 	 {Proceedings of the 37th International Conference on Machine Learning},
  pages = 	 {5338--5348},
  year = 	 {2020},
  volume = 	 {119},
  series = 	 {Proceedings of Machine Learning Research},
  month = 	 {13--18 Jul},
  publisher =    {PMLR},
}

@INPROCEEDINGS{levada_curvature_2022,
  author={Levada, Alexandre L. M.},
  booktitle={2022 26th International Conference on Pattern Recognition (ICPR)}, 
  title={A Curvature based Isometric Feature Mapping}, 
  year={2022},
  volume={},
  number={},
  pages={557-563},
  doi={10.1109/ICPR56361.2022.9956591}}

@article{liu_content-based_2013,
	title = {Content-based image retrieval using color difference histogram},
	volume = {46},
	copyright = {https://www.elsevier.com/tdm/userlicense/1.0/},
	doi = {10.1016/j.patcog.2012.06.001},
	language = {en},
	number = {1},
	urldate = {2024-04-30},
	journal = {Pattern Recognition},
	author = {Liu, Guang-Hai and Yang, Jing-Yu},
	year = {2013},
	pages = {188--198},
}

@inproceedings{liu_swin_2021,
	title = {Swin {Transformer}: {Hierarchical} {Vision} {Transformer} using {Shifted} {Windows}},
	shorttitle = {Swin {Transformer}},
	doi = {10.1109/ICCV48922.2021.00986},
	urldate = {2024-04-04},
	booktitle = {2021 {IEEE}/{CVF} {International} {Conference} on {Computer} {Vision} ({ICCV})},
	author = {Liu, Ze and Lin, Yutong and Cao, Yue and Hu, Han and Wei, Yixuan and Zhang, Zheng and Lin, Stephen and Guo, Baining},
	month = oct,
	year = {2021},
	pages = {9992--10002},
}

@INPROCEEDINGS {liu2022convnet2020s,
author = { Liu, Zhuang and Mao, Hanzi and Wu, Chao-Yuan and Feichtenhofer, Christoph and Darrell, Trevor and Xie, Saining },
booktitle = { 2022 IEEE/CVF Conference on Computer Vision and Pattern Recognition (CVPR) },
title = {{ A ConvNet for the 2020s }},
year = {2022},
volume = {},
pages = {11966-11976},
doi = {10.1109/CVPR52688.2022.01167},
publisher = {IEEE Computer Society},
month =Jun}

@inproceedings{locatello2019challenging,
title	= {Challenging Common Assumptions in the Unsupervised Learning of Disentangled Representations},author	= {Francesco Locatello and Stefan Bauer and Mario Lučić and Gunnar Rätsch and Sylvain Gelly and Bernhard Schölkopf and Olivier Frederic Bachem},year	= {2019},URL	= {http://proceedings.mlr.press/v97/locatello19a.html},note	= {Best Paper Award},booktitle	= {International Conference on Machine Learning}}

@InProceedings{ma2019disengcn,
  title = 	 {Disentangled Graph Convolutional Networks},
  author =       {Ma, Jianxin and Cui, Peng and Kuang, Kun and Wang, Xin and Zhu, Wenwu},
  booktitle = 	 {Proceedings of the 36th International Conference on Machine Learning},
  pages = 	 {4212--4221},
  year = 	 {2019},
  volume = 	 {97},
  series = 	 {Proceedings of Machine Learning Research},
  month = 	 {09--15 Jun},
  publisher =    {PMLR},
  url = 	 {https://proceedings.mlr.press/v97/ma19a.html},
}

@Article{marconato2023interpretability,
AUTHOR = {Marconato, Emanuele and Passerini, Andrea and Teso, Stefano},
TITLE = {Interpretability Is in the Mind of the Beholder: A Causal Framework for Human-Interpretable Representation Learning},
JOURNAL = {Entropy},
VOLUME = {25},
YEAR = {2023},
NUMBER = {12},
ARTICLE-NUMBER = {1574},
URL = {https://www.mdpi.com/1099-4300/25/12/1574},
PubMedID = {38136454},
ISSN = {1099-4300},
DOI = {10.3390/e25121574}
}

@article{mcinnes_umap_2020, 
doi = {10.21105/joss.00861}, 
year = {2018}, 
publisher = {The Open Journal}, 
volume = {3}, 
number = {29}, 
pages = {861}, 
author = {McInnes, Leland and Healy, John and Saul, Nathaniel and Großberger, Lukas}, 
title = {{UMAP:} Uniform Manifold Approximation and Projection}, 
journal = {Journal of Open Source Software} 
}

@inproceedings{nilsback_visual_2006,
	title = {A {Visual} {Vocabulary} for {Flower} {Classification}},
	volume = {2},
	url = {http://ieeexplore.ieee.org/document/1640927/},
	doi = {10.1109/CVPR.2006.42},
	booktitle = {2006 {IEEE} {Computer} {Society} {Conference} on {Computer} {Vision} and {Pattern} {Recognition} ({CVPR})},
	publisher = {IEEE},
	author = {Nilsback, M.-E. and Zisserman, A.},
	year = {2006},
	pages = {1447--1454},
}

@inproceedings{
oikarinen2023labelfreecbm,
title={Label-free Concept Bottleneck Models},
author={Tuomas Oikarinen and Subhro Das and Lam M. Nguyen and Tsui-Wei Weng},
booktitle={The Eleventh International Conference on Learning Representations },
year={2023},
}

@INPROCEEDINGS{parkhi_cats-and-dogs_2012,
  author={Parkhi, Omkar M and Vedaldi, Andrea and Zisserman, Andrew and Jawahar, C. V.},
  booktitle={2012 IEEE Conference on Computer Vision and Pattern Recognition}, 
  title={Cats and dogs}, 
  year={2012},
  volume={},
  number={},
  pages={3498-3505},
  doi={10.1109/CVPR.2012.6248092}}

@Article{pedronette_2021_rdpac,
AUTHOR = {Pedronette, Daniel Carlos Guimarães and Pascotti, Lucas Valem and Latecki, Longin Jan},
TITLE = {Efficient Rank-Based Diffusion Process with Assured Convergence},
JOURNAL = {Journal of Imaging},
VOLUME = {7},
YEAR = {2021},
NUMBER = {3},
ARTICLE-NUMBER = {49},
URL = {https://www.mdpi.com/2313-433X/7/3/49},
PubMedID = {34460705},
ISSN = {2313-433X},
DOI = {10.3390/jimaging7030049}
}

@inproceedings{pedronette_unsupervised_2015,
	title = {Unsupervised {Effectiveness} {Estimation} for {Image} {Retrieval} {Using} {Reciprocal} {Rank} {Information}},
	doi = {10.1109/SIBGRAPI.2015.28},
	booktitle = {2015 28th {SIBGRAPI} {Conference} on {Graphics}, {Patterns} and {Images} ({SIBGRAPI})},
	publisher = {IEEE Computer Society},
	author = {Pedronette, Daniel Carlos Guimaraes and Torres, Ricardo da S.},
	month = aug,
	year = {2015},
	pages = {321--328},
}

@ARTICLE{pedronette_2019_lhrr,
  author={Pedronette, Daniel Carlos Guimarães and Valem, Lucas Pascotti and Almeida, Jurandy and da S. Torres, Ricardo},
  journal={IEEE Transactions on Image Processing}, 
  title={Multimedia Retrieval Through Unsupervised Hypergraph-Based Manifold Ranking}, 
  year={2019},
  volume={28},
  number={12},
  pages={5824-5838},
  doi={10.1109/TIP.2019.2920526}}

@article{PEDRONETTE2021_bfstree,
title = {A BFS-Tree of ranking references for unsupervised manifold learning},
journal = {Pattern Recognition},
volume = {111},
pages = {107666},
year = {2021},
issn = {0031-3203},
doi = {https://doi.org/10.1016/j.patcog.2020.107666},
url = {https://www.sciencedirect.com/science/article/pii/S0031320320304696},
author = {Daniel Carlos Guimarães Pedronette and Lucas Pascotti Valem and Ricardo da S. Torres},
}

@article{pereiraferrero2024unsupervised,
title = {Unsupervised affinity learning based on manifold analysis for image retrieval: A survey},
journal = {Computer Science Review},
volume = {53},
pages = {100657},
year = {2024},
issn = {1574-0137},
doi = {https://doi.org/10.1016/j.cosrev.2024.100657},
author = {V.H. Pereira-Ferrero and T.G. Lewis and L.P. Valem and L.G.P. Ferrero and D.C.G. Pedronette and L.J. Latecki},
}

@inproceedings{perozzi_deepwalk_2014,
	series = {{KDD} '14},
	title = {{DeepWalk}: online learning of social representations},
	doi = {10.1145/2623330.2623732},
	booktitle = {Proceedings of the 20th {ACM} {SIGKDD} {International} {Conference} on {Knowledge} {Discovery} and {Data} {Mining}},
	publisher = {ACM},
	author = {Perozzi, Bryan and Al-Rfou, Rami and Skiena, Steven},
	year = {2014},
	pages = {701--710},
}

@article{piaggesi_dine_2024,
	title = {{DINE}: {Dimensional} {Interpretability} of {Node} {Embeddings}},
	volume = {36},
	doi = {10.1109/TKDE.2024.3425460},
	number = {12},
	journal = {IEEE Transactions on Knowledge and Data Engineering},
	author = {Piaggesi, Simone and Khosla, Megha and Panisson, André and Anand, Avishek},
	year = {2024},
	pages = {7986--7997},
}

@article{
piaggesi2025disentangled,
title={Disentangled and Self-Explainable Node Representation Learning},
author={Simone Piaggesi and Andr{\'e} Panisson and Megha Khosla},
journal={Transactions on Machine Learning Research},
issn={2835-8856},
year={2025},
url={https://openreview.net/forum?id=s51TQ8Eg1e},
note={}
}

@ARTICLE{rudin_stop_2019,
  title     = "Stop explaining black box machine learning models for high
               stakes decisions and use interpretable models instead",
  author    = "Rudin, Cynthia",
  journal   = "Nature Machine Intelligence",
  publisher = "Springer Science and Business Media LLC",
  volume    =  1,
  number    =  5,
  pages     = "206--215",
  month     =  may,
  year      =  2019,
  copyright = "https://www.springernature.com/gp/researchers/text-and-data-mining",
  language  = "en"
}

@INPROCEEDINGS {salehi2019graphattentionautoencoders,
author = { Salehi, Amin and Davulcu, Hasan },
booktitle = { 2020 IEEE 32nd International Conference on Tools with Artificial Intelligence (ICTAI) },
title = {{ Graph Attention Auto-Encoders }},
year = {2020},
volume = {},
ISSN = {},
pages = {989-996},
doi = {10.1109/ICTAI50040.2020.00154},
url = {https://doi.ieeecomputersociety.org/10.1109/ICTAI50040.2020.00154},
publisher = {IEEE Computer Society},
address = {Los Alamitos, CA, USA},
month =Nov}

@inproceedings{tang_line_2015,
	title = {{LINE}: {Large}-scale {Information} {Network} {Embedding}},
	doi = {10.1145/2736277.2741093},
	booktitle = {Proceedings of the 24th {International} {Conference} on {World} {Wide} {Web}},
	publisher = {International World Wide Web Conferences Steering Committee},
	author = {Tang, Jian and Qu, Meng and Wang, Mingzhe and Zhang, Ming and Yan, Jun and Mei, Qiaozhu},
	month = may,
	year = {2015},
	pages = {1067--1077},
}

@INPROCEEDINGS{valem_novel_2022,
  author={Valem, Lucas Pascotti and Atsushi Sato Kawai, Vinicius and Pereira-Ferrero, Vanessa Helena and Carlos Guimarães Pedronette, Daniel},
  booktitle={2022 IEEE International Conference on Image Processing (ICIP)}, 
  title={A Novel Rank Correlation Measure for Manifold Learning on Image Retrieval and Person Re-ID}, 
  year={2022},
  volume={},
  number={},
  pages={1371-1375},
  doi={10.1109/ICIP46576.2022.9898060}}

@article{valem_graph_2023,
	title = {Graph {Convolutional} {Networks} based on manifold learning for semi-supervised image classification},
	volume = {227},
	doi = {10.1016/j.cviu.2022.103618},
	journal = {Computer Vision and Image Understanding},
	author = {Valem, Lucas Pascotti and Pedronette, Daniel Carlos Guimarães and Latecki, Longin Jan},
	year = {2023},
	pages = {103618},
}

@inproceedings{valem2017udlf,
   author = {Valem, Lucas Pascotti and Pedronette, Daniel Carlos Guimar\~{a}es},
   title = {An Unsupervised Distance Learning Framework for Multimedia Retrieval},
   booktitle = {Proceedings of the 2017 ACM on International Conference on Multimedia Retrieval},
   series = {ICMR '17},
   year = {2017},
   isbn = {978-1-4503-4701-3},
   location = {Bucharest, Romania},
   pages = {107--111},
   numpages = {5},
   url = {http://doi.acm.org/10.1145/3078971.3079017},
   doi = {10.1145/3078971.3079017},
   acmid = {3079017},
   publisher = {ACM},
   address = {New York, NY, USA},
}

@inproceedings{vaswani_attention_2023,
 author = {Vaswani, Ashish and Shazeer, Noam and Parmar, Niki and Uszkoreit, Jakob and Jones, Llion and Gomez, Aidan N and Kaiser, \L ukasz and Polosukhin, Illia},
 booktitle = {Advances in Neural Information Processing Systems},
 pages = {},
 publisher = {Curran Associates, Inc.},
 title = {Attention is All you Need},
 volume = {30},
 year = {2017},
}

@inproceedings{
veličković2018deepgraphinfomax,
title={Deep Graph Infomax},
author={Petar Veličković and William Fedus and William L. Hamilton and Pietro Liò and Yoshua Bengio and R Devon Hjelm},
booktitle={International Conference on Learning Representations},
year={2019},
url={https://openreview.net/forum?id=rklz9iAcKQ},
}

@techreport{wah_cub200_2011, 
title={The Caltech-UCSD Birds-200-2011 Dataset}, 
publisher={California Institute of Technology}, 
author={Wah, Catherine and Branson, Steve and Welinder, Peter and Perona, Pietro and Belongie, Serge}, 
year={2011}, 
month={Jul},
url={}}

@INPROCEEDINGS{wang2021tesnet,
  author={Wang, Jiaqi and Liu, Huafeng and Wang, Xinyue and Jing, Liping},
  booktitle={2021 IEEE/CVF International Conference on Computer Vision (ICCV)}, 
  title={Interpretable Image Recognition by Constructing Transparent Embedding Space}, 
  year={2021},
  volume={},
  number={},
  pages={875-884},
  doi={10.1109/ICCV48922.2021.00093}}

@ARTICLE{wang2024disentangled,
  author={Wang, Xin and Chen, Hong and Tang, Si'ao and Wu, Zihao and Zhu, Wenwu},
  journal={IEEE Transactions on Pattern Analysis and Machine Intelligence}, 
  title={Disentangled Representation Learning}, 
  year={2024},
  volume={46},
  number={12},
  pages={9677-9696},
  doi={10.1109/TPAMI.2024.3420937}}

@inproceedings{wu_simplifying_2019,
	series = {Proceedings of {Machine} {Learning} {Research}},
	title = {Simplifying {Graph} {Convolutional} {Networks}},
	volume = {97},
	url = {https://proceedings.mlr.press/v97/wu19e.html},
	booktitle = {Proceedings of the 36th {International} {Conference} on {Machine} {Learning}},
	publisher = {PMLR},
	author = {Wu, Felix and Souza, Amauri and Zhang, Tianyi and Fifty, Christopher and Yu, Tao and Weinberger, Kilian},
	month = jun,
	year = {2019},
	pages = {6861--6871},
}

@INPROCEEDINGS{yang2023LaBo,
  author={Yang, Yue and Panagopoulou, Artemis and Zhou, Shenghao and Jin, Daniel and Callison-Burch, Chris and Yatskar, Mark},
  booktitle={2023 IEEE/CVF Conference on Computer Vision and Pattern Recognition (CVPR)}, 
  title={Language in a Bottle: Language Model Guided Concept Bottlenecks for Interpretable Image Classification}, 
  year={2023},
  volume={},
  number={},
  pages={19187-19197},
  doi={10.1109/CVPR52729.2023.01839}}

@INPROCEEDINGS{zhao2021towards,
  author={Zhao, Wenliang and Rao, Yongming and Wang, Ziyi and Lu, Jiwen and Zhou, Jie},
  booktitle={2021 IEEE/CVF International Conference on Computer Vision (ICCV)}, 
  title={Towards Interpretable Deep Metric Learning with Structural Matching}, 
  year={2021},
  volume={},
  number={},
  pages={9867-9876},
  doi={10.1109/ICCV48922.2021.00974}}

@ARTICLE{zheng2017siftmeetscnndecade,
  author={Zheng, Liang and Yang, Yi and Tian, Qi},
  journal={IEEE Transactions on Pattern Analysis and Machine Intelligence}, 
  title={SIFT Meets CNN: A Decade Survey of Instance Retrieval}, 
  year={2018},
  volume={40},
  number={5},
  pages={1224-1244},
  doi={10.1109/TPAMI.2017.2709749}}

\end{document}